\documentclass[a4paper,fleqn]{cas-dc}

\usepackage[authoryear]{natbib}
\usepackage{wrapfig}
\usepackage{flushend}

\def\tsc#1{\csdef{#1}{\textsc{\lowercase{#1}}\xspace}}
\tsc{WGM}
\tsc{QE}

\begin{document}
\let\WriteBookmarks\relax
\def\floatpagepagefraction{1}
\def\textpagefraction{.001}

\shorttitle{Bi-consolidating Model}

\shortauthors{Xiao Luo et al.}  

\title [mode = title]{A Bi-consolidating Model for Joint Relational Triple Extraction}


%

\author[gz]{Xiaocheng Luo}
\ead{gs.xcluo22@gzu.edu.cn}

\author[gz]{Yanping Chen}
\ead{ypench@gmail.com}
\cormark[1]
\cortext[1]{Corresponding author}

\author[gc]{Ruixue Tang}
\ead{gs.rxtang19@gzu.edu.cn}

\author[gz]{Caiwei Yang}
\ead{gs.cwyang21@gzu.edu.cn}

\author[gz]{Ruizhang Huang}
\ead{rzhuang@gzu.edu.cn}

\author[gz]{Yongbin Qin}
\ead{ybqin@gzu.edu.cn}

\affiliation[gz]{organization={State Key Laboratory of Public Big Data, College of Computer Science and Technology},
	addressline={Guizhou University}, 
	city={Guiyang},
	postcode={550025}, 
	state={Guizhou},
	country={PR China}}

\affiliation[gc]{organization={School of Information},
	addressline={Guizhou University of Finance and Econnomics}, 
	city={Guiyang},
	postcode={550025}, 
	state={Guizhou},
	country={PR China}}



\begin{highlights}
	\item Based on a two-dimensional sentence representation, we propose a bi-consolidating model.
	\item Employing pixel difference convolution to reinforce local semantic features.
	\item Utilizing the mechanism of attention learns remote semantic dependencies in a sentence.
	\item Our model’s effectiveness has been confirmed through experimental results.
\end{highlights}

\begin{abstract} 
Current methods to extract relational triples directly make a prediction based on a possible entity pair in a raw sentence without depending on entity recognition. The task suffers from a serious semantic overlapping problem, in which several relation triples may share one or two entities in a sentence. In this paper, based on a two-dimensional sentence representation, a bi-consolidating model is proposed to address this problem by simultaneously reinforcing the local and global semantic features relevant to a relation triple. This model consists of a local consolidation component and a global consolidation component. The first component uses a pixel difference convolution to enhance semantic information of a possible triple representation from adjacent regions and mitigate noise in neighbouring neighbours. The second component strengthens the triple representation based a channel attention and a spatial attention, which has the advantage to learn remote semantic dependencies in a sentence. They are helpful to improve the performance of both entity identification and relation type classification in relation triple extraction. After evaluated on several publish datasets, the bi-consolidating model achieves competitive performance. Analytical experiments demonstrate the effectiveness of our model for relational triple extraction and give motivation for other natural language processing tasks.
\end{abstract}


\maketitle

\section{Introduction}
\label{sec:Introduction}

Relational triple extraction is defined as a task to identify pairs of entities and extract their relations from unstructured text. The output can be represented in the form of triples: (\emph{subject}, \emph{relation}, \emph{object}) or (s, r, o). Instead of identifying relations from predefined entity types, the relational triple extraction task can find relationships between any entity pairs in a sentence. It has the advantage to find all possible entity relations in a sentence and avoid the cascading failure caused by predicting entity types. In recent years, the task has achieved great attention in natural language processing and is widely adopted to support other downstream tasks such as knowledge graph construction \citep{dong2014knowledge}, information retrieval \citep{sinha2015overview}, etc.

Techniques to support relational triple extraction can be roughly divided into two frameworks: pipeline framework and joint framework. In pipeline framework \citep{chan2011exploiting, zhong2021frustratingly}, the task is divided into two independent subtasks: entity recognition and relation extraction. The main shortcoming is that they suffer from the error propagation problem and ignore the interaction between two subtasks. Recent researches focus on joint framework, which extract the triple (s, r, o) in end-to-end multi-objective models. In addition, to avoid the error propagation, joint models also have the ability to share model parameters between different tasks. They have shown a great potential to support the relational triple extraction task \citep{zheng2017joint, wang2020tplinker, ren2021novel, yu2020jointer}. 

Recent developments in Large Language Models (LLMs) \citep{chowdhery2023palm, touvron2023llama} have demonstrated remarkable capabilities in many NLP applications. As generative language models, they are effective in tasks such as Question Answering \citep{yona2024narrowing} or Summarization \citep{pu2023summarization}. However, in information extraction, they usually suffer from serious performance degeneration. For examples, \cite{han2023information} revealed that LLMs only outperform Small Language Models (SLMs) with limited labels and samples. \cite{ma2023large} highlight that ChatGPT achieved competitive performance only in simpler cases. The unfitness is due to the reason that a sentence usually contains several linguistic units at the same time, especially in the relational triple extraction task. These linguistic units share the same contextual features in a sentence, which leads to a complicated semantic structure. Generative models usually encode a sentence into a dense vector, which easily leads to worse performance when decoding overlapped linguistic units.

\begin{figure}[h] 
	\centering
	\includegraphics[width=\linewidth]{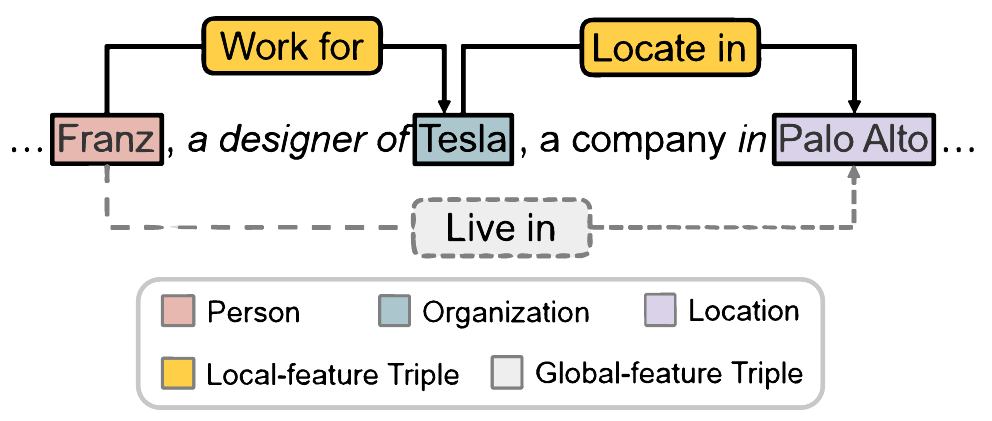}
	\caption{An example of overlapped relational triples.}
	\label{fig:local and global}
\end{figure}

Even great successes have been achieved in this field, relational triple extraction is still a challenging task, because a sentence usually contains several overlapped entity pairs. The overlapping phenomenon can be divided into four categories: Normal, entity pair overlapping (EPO), single entity overlapping (SEO) and subject object overlapping (SOO) \citep{zheng2021prgc}. The Normal refers to relational triples without overlapping. EPO denotes to two triples containing the same two entities but annotated with different relation types. In SEO, two triples share the same subject or object entity. In SOO, both subject and object are simultaneously shared by two relational triples. An example of the overlapping phenomenon is shown in Figure \ref{fig:local and global}.

The above sentence contains three relational triples. The ``\emph{Live in}'' relation and ``\emph{Locate in}'' relation share the same entity ``\emph{Palo Alto}'', which leads to the SEO overlap. On the other hand, the sentence also contains a SOO overlap between the ``\emph{Locate in}'' relation and ``\emph{Work for}'' relation. The entity ``\emph{Tesla}'' acts as the subject and object in the two relational triples. Because overlapped relational triples share the same contextual features, it is difficult to distinguish them, especially in a sequential sentence representation, where a relational triple is usually represented by fusing two entity representations.

Recent works have also addressed the semantic overlap problem in relational triple extraction, such as SOIRP \citep{dai2024soirp} and RTF \citep{an2024rtf}. These methods address the semantic overlap through a tagging scheme or by enhancing specific semantic representations. Compared with them, our method utilizes a pixel difference convolution and an attention mechanism to simultaneously reinforce both local and global semantic features relevant to a relation triple. By integrating local and global semantic features related to a relation triple, our approach is effective to enrich the overall semantic representation of relational triples.

Additionally, there is a trend in recent works to transform a sentence into a two-dimensional (2D) sentence representation, e.g., the planarized sentence representation \citep{geng2023planarized} or the table filling \citep{li2022unified}. In the 2D representation, each element denotes an entity pair representation in a sentence. Because the 2D sentence representation unfolds a semantic plane, it is effective to represent all entity pairs in a sentence. Based on the 2D representation, the relational triple extraction is conducted as element classification. The main problem is that entity semantics are spread out in the 2D representation, which is influential on the relational triple extraction. Before to show the motivation of our model, we first give a 2D sentence representation in Figure \ref{fig:local and global}.

\begin{figure}[h!]
	\centering
	\normalsize
	\caption{2D Sentence Representation}
	\small
	\vspace{0.1cm}
	\begin{minipage}[t]{0.49\linewidth}
		\label{fig:sent_rep_a}
		\centering
		\includegraphics[width=\linewidth]{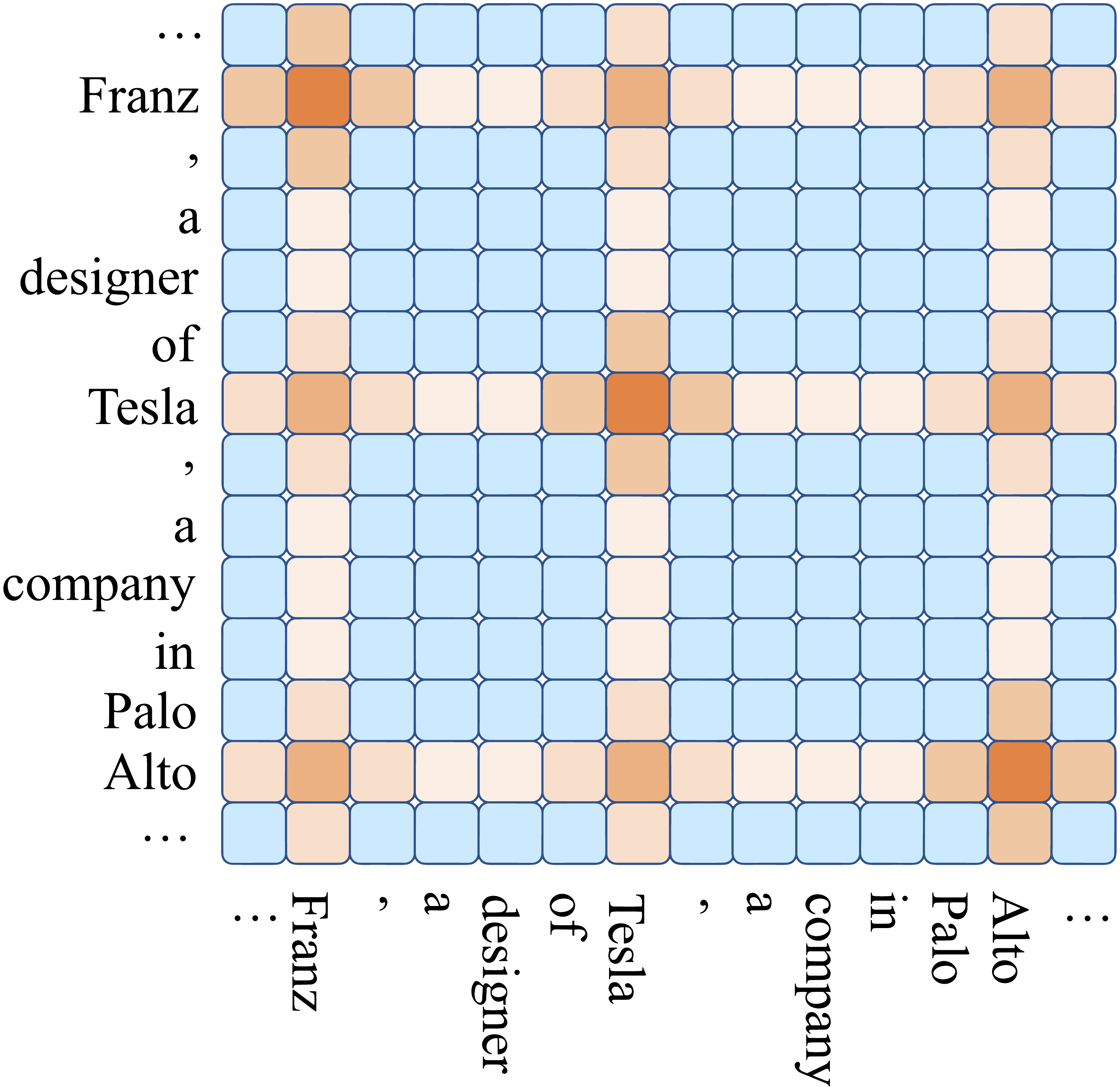}
		\centerline{\hspace{4ex} $(a)$ Theory Representation}
	\end{minipage}
	\begin{minipage}[t]{0.48\linewidth}
		\label{fig:sent_rep_b}
		\centering
		\includegraphics[width=\linewidth]{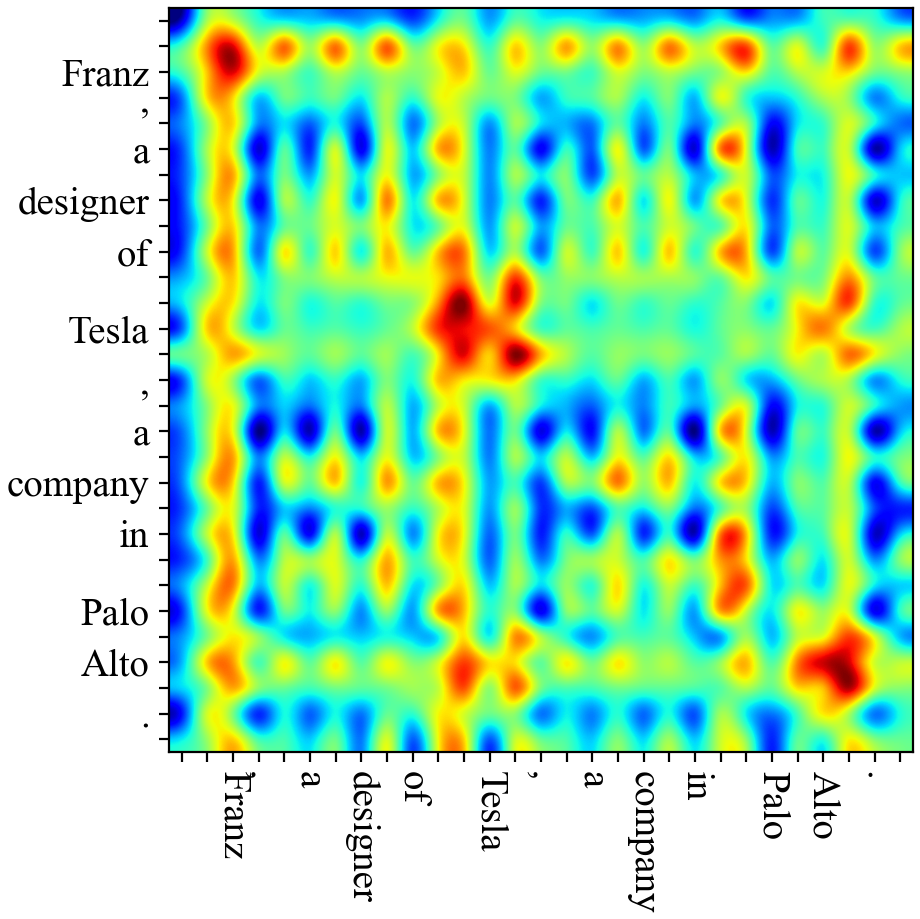}
		\centerline{\hspace{4ex} $(b)$ Real Representation}
	\end{minipage}
	\label{fig:sent_rep}
\end{figure}

Figure \ref{fig:sent_rep} contains two 2D representations about the same sentence in Figure \ref{fig:local and global}, where deep colors represent the strength of the entity semantic in an element (also known as ``span''). Sub-figure (a) is a theory sentence representation. It indicates that, in the 2D sentence representation, the semantics of entities (e.g., ``\emph{Franz}'', ``\emph{Tesla}'' and ``\emph{Alto}'') are shared by all elements in the same columns and rows. Sub-figure (b) is a real sentence representation. To generate this representation, the sentence is first fed into a well trained model, which was conducted with the settings in Section \ref{Experimental Settings}. Then, the real representation is visualized by using the 2D sentence representation in the trained model.

In Figure \ref{fig:sent_rep}, the theory and real representation show the same semantic pattern, which indicates four important phenomena in relational triple extraction. First, the semantic plane is usually constructed by concatenating two word representations. Adjacent elements denote to overlapped phrases which share the same contextual features. Second, two entities of a relational triple can be separated by great distances, e.g., the triple  ( ``\emph{Franz}'',  ``\emph{Live in}'', ``\emph{Palo Alto}''). Third, in the 2D representation, the semantic information of a real entity is shared by all elements in the same column or the same row, because these elements are generated from the entity representation. Fourth, a sentence often contains several relation triples which may share one or two entities. Therefore, the relation triple extraction task suffers from serious semantic overlapping problems in the 2D sentence representation.

Inspired by the aforementioned analyses, in this paper, we propose a bi-consolidating model to address the mentioned problems. This model consists of a local consolidation component and a global consolidation component. The first component uses a pixel difference convolution to enhance the semantic information of a possible triple representation from adjacent regions and mitigate noise in neighbouring neighbours. The second component strengthens the triple representation based a channel attention and a spatial attention, which has the advantage to learn remote semantic dependencies in a sentence. The bi-consolidating model simultaneously reinforce the local and global semantic features relevant to each relation triple. They are helpful to improve the performance of both entity identification and relation type classification in relation triple extraction. To prove the effectiveness of our model, we evaluate the proposed model on four benchmark datasets: $\mathrm{NYT10}$, $\mathrm{NYT11}$, $\mathrm{NYT}$ and $\mathrm{WebNLG}$. Extensive experiments show that it achieves competitive performance and demonstrates the effectiveness of our model for relational triple extraction. In summary, our main contributions are as follows:

\begin{itemize}
	\item[1)] Based on a 2D sentence representation, we propose a bi-consolidating model for extracting relational triples, which has the ability to simultaneously reinforce the local and global semantic features of a relational triple. Compared with related works, we achieved competitive performance on several public evaluation datasets.
	\item[2)] Several analytical experiments are conducted to demonstrate the effectiveness of our model. The results show details of the bi-consolidating model of the 2D sentence representation. It also gives motivations for researches in the relational triple extraction and other NLP tasks.
\end{itemize}

The rest of this paper is organized as follows. Section \ref{sec:related_work} provides an introduction to related works on joint entity and relation extraction. In Section \ref{sec:methodology}, the bi-consolidating model is presented and in details. The effectiveness of our model for joint entity and relation extraction is demonstrated through experiments conducted in Section \ref{sec:experiment}. Furthermore, in Section \ref{sec:analyis}, different components are evaluated to show the effectiveness of the proposed models in relation triple extraction. Finally, Section \ref{sec:conclusion} offers a conclusion and outlines future works.

\section{Related Work}
\label{sec:related_work}

Early studies \citep{zelenko2003kernel, zhou2005exploring, chan2011exploiting} often use a pipeline framework to support the joint entity and relation extraction task. They first identify named entities in the input texts. Then, the relation between all entity pairs are verified for extracting relations between them. Pipeline approaches have two shortcomings in entity and relation extraction. Firstly, it breaks the interaction between entity recognition and relation prediction. Secondly, the relation extraction task suffers from the error propagation caused in falsely named entities.

To address these limitations, researchers have delved into joint extraction methods \citep{sui2023joint, ning2023od, zhang2023rs}, which simultaneously extract entities and relations in a unified framework. This paper mainly focuses on joint methods. In the followings, we roughly divided them into three categories: tagging based methods, Seq2Seq based methods and table filling based methods. Each category is introduced as follows.

\subsection{Tagging based Methods}

Tagging based methods convert the process of triple extraction into a series of interdependent sequence labeling tasks. For example, \cite{wei2020novel} have proposed a CasRel model, which detects all potential head entities and subsequently applies a relation-specific sequence tagger to identify their corresponding tail entities. \cite{zhan2022simple} presented a comprehensive approach, known as DropRel, to tag the triple boundaries and relation types via three distinct dependence relations. Furthermore, to ensure the preservation of semantic information while generating distinct vectors for a given token pair, they proposed a dropout-normalization layer. Recently, \cite{Wang2023CLFMCL} proposed a relation-specific sequence tagging decoder that employs a filter-attention mechanism to highlight more informative features.
However, it is worth noting that methods based on sequence tagging often suffer from issues related to semantic loss and error propagation.

\subsection{Seq2Seq based Methods}

Seq2Seq based methods treat a triple as a sequence of tokens and utilize an encoder-decoder framework to produce relational triples. For example, CopyRE \citep{zeng2018extracting} utilized as a copy mechanism to generate the relationship between two corresponding entities. The main problem with this model is that it is weak to extract the order of relational facts within a sentence. In order to address this limitation, \cite{zeng2019learning} introduced a reinforcement learning to improve this CopyRE model. However, it is only capable of predicting the last word of an entity. In order to overcome this constraint, CopyMTL \citep{zeng2020copymtl} employed as a multi-task learning framework to address the multi-token entity problem. TDEER \citep{li2021tdeer} proposed a novel translating decoding schema that represents the relation as a translation operation from subjects to objects. Even the Seq2Seq-based approaches have great potential to support relation extraction, it still poses challenges in constructing a vector with rich semantic features, particularly for long sentences.

\subsection{Table Filling based Methods}

Table filling based methods formulate the task of triple extraction as a table filling process. For example, SPTree \citep{miwa2016end} effectively captured both word sequence and dependency tree substructure information. This is accomplished through the use of stacked bidirectional tree-structured LSTM-RNNs on top of bidirectional sequential LSTM-RNNs. TF-MTRNN \citep{gupta2016table} proposed a novel context-aware joint entity and word-level relation extraction approach through semantic composition of words. This model is capable of modeling multiple relation instances without knowing the corresponding relation arguments in a sentence. RS-TTS \citep{zhang2023rs} predicted potential relations to avoid computational redundancy and utilizes an efficient tagging and scoring strategy for entity decoding.

In addition to the above mentioned methods, researchers have investigated other approaches. For example, \cite{bekoulis2018joint} presented a multi-head model, which formulates the relational triple extraction task as a multi-head selection problem. \cite{li2019entity} implemented the task as a multi-turn question answering task, where a multi-turn QA model was proposed. \cite{fu2019graphrel} employed a graph convolutional network-based method known as Graphrel. \cite{eberts2020span} utilized a span extraction approach. \cite{sun2021progressive} proposed a multi-task learning-based RTE model. In the field of attention, \cite{xu2015show} proposed a visual attention method to model the importance of features in the image caption task. A residual attention network \citep{wang2017residual} was proposed with a spatial attention mechanism using downsampling and upsampling. Besides, a channel attention mechanism was proposed by SENet \citep{hu2018squeeze}.

\section{Methodology}
\label{sec:methodology}

\begin{figure*}[h]
	\centering
	\includegraphics[width=1\linewidth]{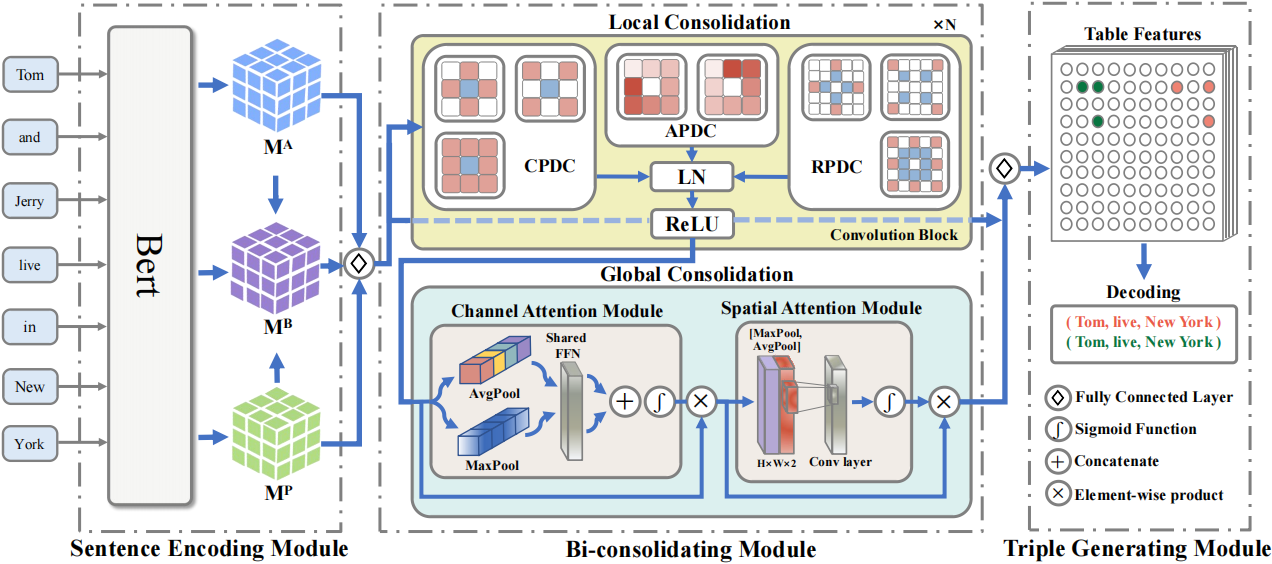}
	\caption{Model Architecture. In the convolution block, the rounded rectangles represent the optional convolution type. The dashed line indicates the residual connections of the model.}
	\label{fig:model}
\end{figure*}

The architecture of the bi-consolidating model is demonstrated in Figure \ref{fig:model}. It comprises three modules: \emph{Sentence Encoding Module}, \emph{Bi-consolidating Module} and \emph{Triple Generating Module}. 

The sentence encoding module transforms a sentence into a two-dimensional representation. In the bi-consolidating module, two components are designed to reinforce the local and global semantic features, respectively. Finally, the triple generating module verifies and outputs extracted relational triples. Each module is discussed as follows.

\subsection{Sentence Encoding Module}
\label{Sentence Encoding}

Given a sentence consisting of $N$ words, it can be denoted as $\mathbf{S} = [x_1,x_2, \ldots, x_N]$ or $\mathbf{S} = x_{[1:N]}$. To capture the semantic information of each word, a common approach is to utilize a pretrained language model that maps each word into an abstract representation. In our experiments, each sentence is fed into the BERT \citep{devlin2019bert} for learning the high-order abstract semantic representations of words. The process can be formalized as:
\begin{equation}
	\begin{aligned}
		[h_1, h_2, \ldots, h_{\mathrm{N}}] & = \text{BERT}\left([x_1, x_2, \ldots, x_N]\right) \\
	\text{or},\ \	\mathbf{H} & = \text{BERT}\left( \mathbf{S} \right)
	\end{aligned}
\end{equation}

\noindent where $\mathbf{H} \in \mathbb{R}^{N \times D_h}$ is a sequence of token representations. The dimension of $h_i$ is denoted as $D_h$.

We adopt a self-cross embedding \citep{geng2023planarized} to map the sequence $\mathbf{H} $ into a 2D sentence representation, represented as a matrix $\mathbf{M}^{B}$. It is represented as:
\begin{equation}
	\begin{aligned}
		\mathbf{M}^{B} = \emph{Crossing} (\mathbf{H} , \mathbf{H} ) \\
	\end{aligned}
\end{equation}

\noindent where, $\mathbf{M}^{B}_{ij} = h_i \oplus h_j $. The symbol “$\oplus$” denotes the concatenation operation. The element $\mathbf{M}^{B}_{ij}$ is a span representation of $[x_i, \ldots , x_j]$ (or $x_{[i:j]}$) in $\mathbf{S}$, denoted as a vector with $2D_h$ dimensions. Therefore, $\mathbf{M}^{B}$ is a $\mathbb{R}^{N \times N \times 2D_h}$ matrix.

To learn structural features of a sentence, following \cite{geng2023planarized} and \cite{li2022unified}, the positional embedding and attention embedding are applied to generate a position matrix($\mathbf{M}^P$) and an attention matrix ($\mathbf{M}^A$) encoding semantic dependencies between elements. The two embedding are represented as follows:
\begin{equation}
\label{equ:embedding}
	\begin{aligned}
		\mathbf{M}^P & = \text{P-Embedding} (P) \\
		\mathbf{M}^A & = \text{A-Embedding} (S) \\
	\end{aligned}
\end{equation}

\noindent where $P_{ij} = N - i$ if $j > i$, else $P_{ij} = j - N$. Therefore, $\mathbf{M}^P$ and $\mathbf{M}^A$ are a $\mathbb{R}^{N \times N \times D_p}$ matrix and $\mathbb{R}^{N \times N \times D_a}$ matrix, respectively.

Finally, the 2D sentence representation is generated as:
\begin{equation}
		\mathbf{M}^{so} = \sigma (W_{so}(\mathbf{M}^{P}  \oplus  \mathbf{M}^{A} \oplus  \mathbf{M}^{B}) + b_{so})
\end{equation}

\noindent where $W_{so} \in \mathbb{R}^{(D_p + D_a + 2D_h) \times D_h}$ and $b_{so} \in \mathbb{R}^{D_h}$ are the trainable parameters. $\sigma$ is the ELU activation function.

\subsection{Bi-consolidating Module}
\label{Semantic Features Learning}

The quality of entity representations and relation representations are influential on the performance of the relational triple extraction. Entity representations heavily depend on the local contextual features present in a sentence. On the other hand, relations encompass semantic expressions that pertain to the entirety of a sentence. Therefore, the global features are more important for relation extraction.

To utilize local and global features,  we design a novel architecture to advance the discriminability of a neural network, which is composed of a local consolidation component and a global consolidation component. Specifically, a pixel difference convolution is introduced to enhance the local information, which is implemented in adjacent regions of neighbours, thereby reinforcing local semantic features. To strengthen global semantic features, we incorporate a channel attention and a spatial attention to capture remote semantic dependencies. The local consolidation and global consolidation are discussed as follows.

\subsubsection{Local Consolidation}
\label{sec:local_consolidation}

The local consolidation component consists of a stack of multiple convolution blocks. A convolution block comprises two layers: a pixel difference convolution layer and a normalization layer \citep{ba2016layer}. The local consolidation component is constructed by stacking ${L}$ identical convolution blocks. The default value of ${L}$ is four in our experiments. The pixel difference convolution and layer normalization layer are introduced as follows:

\textbf{Pixel Difference Convolution}: the process of pixel difference convolution (PDC) is represented in Equation (\ref{equ:pdc_vc}), where the convolution kernels operate on a local patch and replace the original pixels with pixel differences.
\begin{equation}
\label{equ:pdc_vc}
	\begin{aligned}
		& u=f(\nabla \boldsymbol{z}, \boldsymbol{\theta})=\sum_{\left(z_i, z_i^{\prime}\right) \in \mathcal{P}} w_i \cdot\left(z_i-z_i^{\prime}\right) 
	\end{aligned}
\end{equation}

\noindent where, $z_i$ and $z_i^{\prime}$ are the input pixels, which denote to elements in the 2D sentence representation ($\mathbf{M}^{so} $). $w_i$ is the weight of the $k \times k$ convolution kernel. $\mathcal{P}=$ $\{ (z_1, z_1^{\prime}), (z_2, z_2^{\prime})$ $ ,\ldots,(z_m, z_m^{\prime})\}$ ($m \leq k \times k$) is the set of pixel pairs picked from a local patch in $\mathbf{M}^{so}$. 

\begin{figure}[h]
	\centering          
	\includegraphics[width=\linewidth]{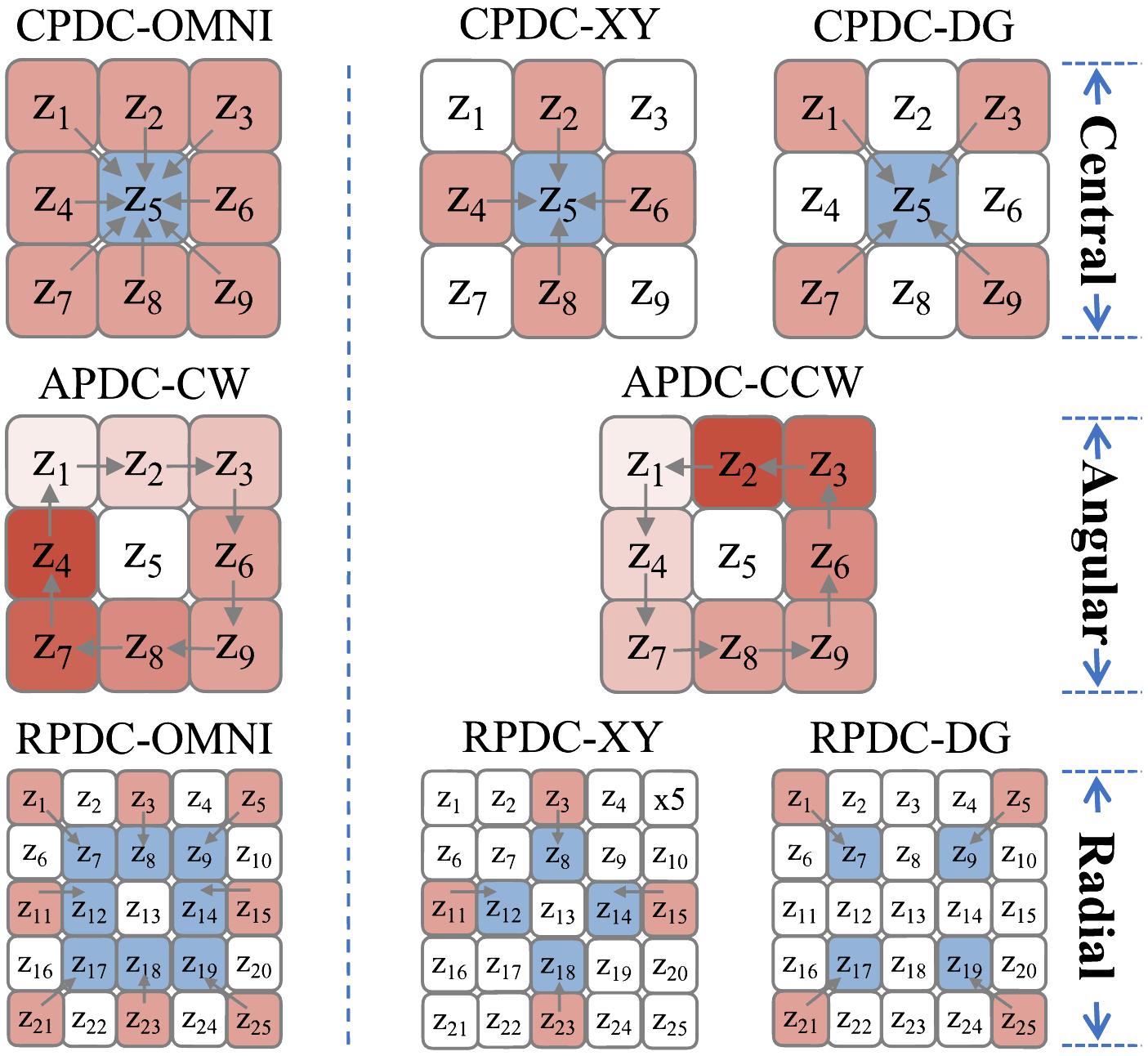}
	\caption{\label{fig:pdc} Eight strategies for selecting pixel pairs are created.}
\end{figure}

Various strategies can be employed to select pixel pairs from the representation $\mathbf{M}^{so}$. In our study, we adopt three kinds of pixel difference convolution presented in \cite{su2021pixel}, named as: central PDC (CPDC), angular PDC (APDC) and radial PDC (RPDC). We also propose five new strategies to encode pixel relations from different directions (XY-axis, diagonal, omnidirectional and bidirectional) to improve the ability for capturing the local dependencies between neighbouring elements. The strategies to select pixel pairs are as illustrated in Figure \ref{fig:pdc}, in which each patch denotes to a convolution kernel. The arrows between elements represent a pixel pair from $(z_i, z_i^{\prime})$, which indicate the direction of semantic dependencies between them.

As shown in Figure \ref{fig:pdc}, in the first row, the CPDC-based methods focus on extracting features from central and adjacent spatial positions. They are enhanced through the use of three coding directions: CPDC-XY for the XY axis, CPDC-DG for diagonal directions, and CPDC-OMNI for omnidirectional directions. In the second row, the APDC-based methods leverage semantic contextual information from neighboring elements to aid in entity boundary detection. They have two bidirectional coding directions: APDC-CW for the clockwise and APDC-CCW for the anticlockwise. In the third row, the RPDC-based methods extract informative features from the peripheral regions. They are also accompanied by three coding directions: RPDC-XY for the XY axis, RPDC-DG for diagonal directions, and RPDC-OMNI for omnidirectional directions. 

To explain the notion of pixel pairs in each local patch, we introduce the APDC-CW patch as an example. As showing in the $1$-th column of the $2$-th row, the APDC-CW utilizes a kernel size of $3 \times 3$. A total of 8 pixel pairs is constructed in the angular clockwise direction within the local patch. Using Equation (\ref{equ:pdc_vc}), the pixel differences are derived from these paired elements, then fed into a convolution layer, which entails element-wise multiplication with kernel weights followed by the summation.

\textbf{Layer Normalization}: when the number of convolution block is increased during the optimization process, the gradient descent method will change the distribution of input features. To ensure the stability of these feature distributions, a layer normalization layer is employed to normalize the output of the pixel difference convolution. This normalization process enables a larger learning rate and accelerates the convergence of the model. Moreover, it is helpful to mitigate the overfitting problem, making the training process smoother. 

\subsubsection{Global Consolidation}
\label{sec:global_consolidation}

The global consolidation component is set to facilitate the model's capture of global information from a 2D sentence representation. As shown in Figure \ref{fig:model}, the global consolidation component comprises a channel attention module and a spatial attention module. The channel attention is applied globally, whereas the spatial attention operates on a local scale. Therefore, channel attention and spatial attention are complementary to each other. Spatial attention highlights keywords related to entities or relations in a given sentence, while channel attention captures global semantic features between entities and relations.

\textbf{Channel Attention Module}: we first provide a mathematical description of the inverse 2D Discrete Cosine Transform (DCT). It is expressed as follows:
\begin{equation}
	\begin{aligned}
		\mathrm{A}_{\mathrm{i}, \mathrm{j}}^{2d} & =\sum_{\mathrm{h}=0}^{\mathrm{H}-1} \sum_{\mathrm{w}=0}^{\mathrm{W}-1} \mathrm{f}_{\mathrm{h}, \mathrm{w}}^{2d} \underbrace{\cos \left(\frac{\pi \mathrm{h}}{\mathrm{H}}\left(\mathrm{i}+\frac{1}{2}\right)\right) \cos \left(\frac{\pi \mathrm{w}}{\mathrm{W}}\left(\mathrm{j}+\frac{1}{2}\right)\right)}_{\text{DCT weights}} \\
		& = \mathrm{f}_{0,0}^{2d} \mathrm{~B}_{0,0}^{\mathrm{i}, \mathrm{j}}+\mathrm{f}_{0,1}^{2d} \mathrm{~B}_{0,1}^{\mathrm{i}, \mathrm{j}}+\cdots+\mathrm{f}_{\mathrm{H}-1, \mathrm{~W}-1}^{2d} \mathrm{~B}_{\mathrm{H}-1, \mathrm{~W}-1}^{\mathrm{i}, \mathrm{j}} \\
		& = \underbrace{\operatorname{GAP}(\mathrm{X}) \mathrm{HW} \mathrm{B}_{0,0}^{\mathrm{i}, \mathrm{j}}}_{\text {utilized }}+\underbrace{\mathrm{f}_{0,1}^{2d} \mathrm{~B}_{0,1}^{\mathrm{i}, \mathrm{j}}+\cdots+\mathrm{f}_{\mathrm{H}-1, \mathrm{~W}-1}^{2d} \mathrm{~B}_{\mathrm{H}-1, \mathrm{~W}-1}^{\mathrm{i}, \mathrm{j}}}_{\text {discarded }}\label{equ:the inverse 2D DCT} \\
		& \quad \text { s.t. } i \in\{0,1, \cdots, \mathrm{~H}-1\}, j \in\{0,1, \cdots, \mathrm{~W}-1\}
	\end{aligned}
\end{equation}

\noindent where DCT weights are abbreviated as $\mathrm{B}^{\mathrm{i}, \mathrm{j}}_{\mathrm{h}, \mathrm{w}}$. $\mathrm{f}^{2d} \in \mathbb{R}^{\mathrm{H} \times \mathrm{W}}$ is the 2D DCT frequency spectrum. $\mathrm{H}$ is the height of $\mathrm{A}^{2d}$. $\mathrm{W}$ is the width of $\mathrm{A}^{2d}$.

It can be concluded from the above equation that the previous application of channel attention solely focused on the lowest frequency component of the first term GAP (Global Average Pooling). It ignores the subsequent components expressed in Equation (\ref{equ:the inverse 2D DCT}), resulting in disregarded information. However, the excessive utilization of frequency components leads to the generation of superfluous information. It is imperative to take into account the optimal selection of frequency components. Therefore, in our work, we adopt an average and a max pooling operation to improve the performance. 

The averaging pooling serves to eliminate redundant information while concurrently retaining the most significant information within a sentence. In comparison, max pooling enables the identification and focuses on the most salient features presented within a sentence.  Spatial information from feature maps is aggregated through the average and max pooling operations. It generates two distinct spatial context descriptors: $\mathbf{U}^\mathbf{c}_{\mathbf{avg}}$, representing the average-pooled features, and $\mathbf{U}^\mathbf{c}_{\mathbf{max}}$, representing the max-pooled features.

Subsequently, these descriptors ($\mathbf{U}^\mathbf{c}_{\mathbf{avg}}$ and $\mathbf{U}^\mathbf{c}_{\mathbf{max}}$) are fed into a shared feedforward neural network (FFN), followed by the utilization of a sigmoid function to generate channel attention weights $\mathbf{Q}_\mathbf{c} \in \mathbb{R}^{C \times 1 \times 1}$. The channel attention is computed as:
\begin{equation}
	\begin{aligned}
		\mathbf{Q}_{\mathbf{c}} & =\sigma\left(\emph{FFN}\left(\emph{AvgPool}(\mathbf{U})\right) + \emph{FFN}\left(\emph{MaxPool}(\mathbf{U})\right)\right) \\
		& =\sigma\left(\mathbf{U}_{\mathbf{a v g}}^{\mathbf{c}}\mathbf{W_c} + \mathbf{U}_{\mathbf{m a x}}^{\mathbf{c}}\mathbf{W_c}\right)
	\end{aligned}
\end{equation}

\noindent where $\sigma$ denotes the sigmoid function, $\mathbf{W_c} \in \mathbb{R}^{C \times C}$. Note that the FFN weights $\mathbf{W_c}$ is shared for both inputs.

The shared feedforward neural network is designed to capture the cross-channel interaction, which distinguishes them from the traditional fully-connected (FC) layers with dimensionality reduction. Our ablation studies (Section \ref{Ablations of model components}) demonstrate that dimensionality reduction results in adverse effects on channel attention prediction. Therefore, reducing the number of parameters is unnecessary.

\textbf{Spatial Attention Module}: in contrast to channel attention focuses on learning across channels to emphasize relevant information, spatial attention involves learning across spatial dimensions to highlight important regions within an input. These two mechanisms provide a comprehensive approach to capturing the significance of different aspects within the data. Similar to channel attention, we aggregate channel information from feature maps through the utilization of two pooling operations, resulting in the generation of two 2D maps:  $\mathbf{U}^\mathbf{s}_{\mathbf{max}} $ and $\mathbf{U}^\mathbf{s}_{\mathbf{avg}}$. Each map represents the average and max pooling features of each channel, respectively. These maps are then concatenated and subjected to convolution through a convolution configured as ``CNN-2D'', thereby generating our 2D spatial attention map $\mathbf{Q}_\mathbf{s} \in \mathbb{R}^{H \times W}$. The spatial attention is computed as:
\begin{equation}
	\begin{aligned}
		\mathbf{Q}_{\mathbf{s}} & =\sigma\left({F}_s([\emph{AvgPool}(\mathbf{U}) \oplus \emph{MaxPool}(\mathbf{U})])\right) \\
		& =\sigma\left({F}_s\left(\left[\mathbf{U}_{\mathbf{a v g}}^{\mathbf{s}} \oplus \mathbf{U}_{\mathbf{max} }^{\mathbf{s}}\right]\right)\right)
	\end{aligned}
\end{equation}

\noindent where $\sigma$ represents the sigmoid function and ${F}_s$ denotes a convolution operation using a filter size of $7 \times 7$.

The representation ($\mathbf{M}^{so}_{i, j}$) is simultaneously fed into the local consolidation component and the global consolidation component to learn local and global semantic features. The process is represented as:
\begin{equation}
	\mathbf{T}^{so}_{i, j} = \operatorname{Attention}\left(\left[\operatorname{LN}(\operatorname{Conv^{\varepsilon}}(\mathbf{M}^{so}_{i, j}))\right]_{\times 4} \right)  \label{equ:T_so}
\end{equation}

\noindent where $\varepsilon \in \{\text{CPDC}, \text{APDC}, \text{RPDC}, \text{CNN-2D}\}$ denotes the type of convolution. The number ``4'' indicates the number of convolution block stacked in the convolution component. The output $\mathbf{T}^{so}$ is 2D sentence representation, which encodes the subject and object features relevant to all possible named entity pairs in a sentence. 

\subsection{Triple Generating Module}
\label{Triple Generating}

In this module, the 2D sentence representation $\mathbf{T}^{so}$ is fed into a two-layer perceptron to learn a table feature for each relation candidate. The table feature for a relation $r$ is denoted as $\mathbf{T}^{r}$. Each item in $\mathbf{T}^{r}$ represents the label feature for a token pair. Specifically, for a token pair $(x_i, x_j)$, we represent its label feature as $\mathbf{T}^{r}(i, j)$, which is computed by the following Equation (\ref{equ:classifier}).
\begin{equation}
	\mathbf{T}^{r}(i, j) = W_r \operatorname{ReLU}(drop(\mathbf{T}^{so}_{i, j}W + b)) + b_r  \label{equ:classifier}
\end{equation}

\noindent where $W \in \mathbb{R}^{D_h \times d}$, $b$ are trainable weight and bias. $W_r \in \mathbb{R}^{d \times K \times |Y|}$, $b_r$ are trainable weight and bias. Here, $K$ signifies the number of predefined relations, $|Y|$ denotes the number of predicted labels. The function $drop(\cdot)$ denotes the dropout strategy \citep{srivastava2014dropout} utilized for mitigating overfitting. 

Next, all relation triples can be generated by utilizing the table features $\mathbf{T} \in \mathbb{R}^{N \times N \times K \times |Y|}$. Specifically, for every relation, the method outlined in Equation (\ref{equ:table}) is employed to populate its corresponding table.
\begin{equation}
	\begin{aligned}
		& \hat{\operatorname{table}_r}(i, j)=\operatorname{softmax}\left(\mathbf{T}^r(i, j)\right) \\
		& {\operatorname{table}_r}(i, j)=\underset{y \in Y}{\operatorname{argmax}}\left(\hat{\operatorname{table}_r}(i, j)[y]\right)
		\label{equ:table}
	\end{aligned}
\end{equation}

\noindent where $ \hat{table_r}(i, j) \in \mathbb{R}^{|Y|}$, $table_r(i, j)$ is the labeled result for the token pair $(x_i, x_j)$ in the table of the relation $r$, and \emph{$y \in \{$``N/A'', ``B-B'', ``B-E'', ``E-E''$\}$}. ``B'' and ``E'' denote to the beginning token and end token, respectively. Therefore, when \emph{$y$ = ``B-B''}, it indicates the beginning tokens of both the subject and the object. Similarly, when \emph{$y$ = ``B-E''} or \emph{$y$ = ``E-E''}, it represents the beginning and end tokens of either the subject or the object. On the other hand, if \emph{$y$ = ``N/A''}, it means that the pair $(x_i, x_j)$ does not conform to any of the aforementioned cases.

Finally, the filled tables are decoded. All triples are deduced using the following method. The subject can be obtained through splicing from ``B-E'' to ``E-E''. The object can be obtained through splicing from ``B-B'' to ``B-E''. They result in the acquisition of the relational triple. It is noteworthy that the two paired entities share a common ``B-E''.

We define the model the objective function as follows:

\begin{equation}
	\begin{aligned}
		\mathcal{L}_{\text {triple }}= & -\frac{1}{N \times K \times N} \times \\
		& \sum_{i=1}^N \sum_{k=1}^K \sum_{j=1}^N \log \left (\operatorname{table}_r(i, j) = \operatorname{gold}_r(i, j)\right)
		\label{}
	\end{aligned}
\end{equation}

\noindent where $\operatorname{gold}_r(i, j)$ denotes the gold tag obtained from annotations. The codes to implement our bi-consolidating model is available at: \href{https://github.com/xiaocen-luo/bi-consolidating-model}{https://github.com/xiaocen-luo/bi-consolidating-model}.

\section{Experiment}
\label{sec:experiment}

\subsection{Experimental Settings}
\label{Experimental Settings}

\begin{table*}[h]
	\caption{\label{table:datasets} Statistics of the datasets, where $I$ denotes the number of triples within a sentence. Note that a sentence may simultaneously hold patterns of EPO, SEO and SOO.}
	\resizebox{\linewidth}{!} {
		\begin{tabular}{l ccccc cccccccccc}
			\toprule[2pt]
			\multirow{2}{*}{Category} & \multicolumn{4}{c}{Dataset}&\multicolumn{11}{c}{Details of Test Set}\\
			\cline{2-5}  \cline{7-16}  
			& \multicolumn{1}{c}{Train} & Valid & Test & Relation & & Normal & SEO & EPO & SOO & $I=1$ & $I=2$ & $I=3$ & $I=4$ & $I \geq 5$ & Triples\\
			\hline
			$\mathrm{NYT}^*$& 56,195& 4,999& 5,000& 24& & 3,266& 1,297& 978& 45& 3,244& 1,045& 312& 291& 108& 8,110 \\
			$\mathrm{WebNLG}^*$& 5,019& 500& 703& 171& & 245& 457& 26& 84& 266& 171& 131& 90& 45& 1,591 \\
			$\mathrm{NYT}$& 56,195& 5,000& 5,000& 24& & 3,222& 1,273& 969& 117& 3,240& 1,047& 314& 290& 109& 8,120 \\
			$\mathrm{WebNLG}$& 5,019& 500& 703& 216& & 239& 448& 6& 85& 256& 175& 138& 93& 41& 1,607 \\
			$\mathrm{NYT10}$& 69988& 351& 4006& 29& & 2813& 742& 715& 185& 2950& 595& 187& 239& 35& 5859 \\
			$\mathrm{NYT11}$& 62335& 313& 369& 12& & 363& 1& -& 5& 368& 1& -& -& -& 370 \\
			\bottomrule[2pt]
		\end{tabular}
	}
\end{table*}

\textbf{Datasets}: we evaluate the bi-consolidating model on four benchmark datasets: $\mathrm{NYT10}$ \citep{riedel2010modeling}, $\mathrm{NYT11}$ \citep{hoffmann2011knowledge}, $\mathrm{NYT}$ \citep{riedel2010modeling} and $\mathrm{WebNLG}$ \citep{gardent2017creating}. 
In order to conduct a fair comparison with related works, we employ settings following the latest works \citep{ren2022simple,sun2021progressive, wang2020tplinker}. 
They utilize the NYT and WebNLG datasets provided by \cite{zeng2018extracting}, and the NYT10 and NYT11 datasets published by \cite{takanobu2019hierarchical}. 

In these datasets, both NYT and WebNLG have two distinct versions. 
Specifically, $\mathrm{NYT}^*$ and $\mathrm{WebNLG}^*$ are annotated with the tail tokens of a given entity. $\mathrm{NYT}$ and $\mathrm{WebNLG}$ are annotated with the whole range of an entity. Table \ref{table:datasets} presents the statistical information of these datasets.

To evaluate the ability of our model for detecting overlapped relation instances, we follow the \cite{zeng2018extracting} to partition the test set based on patterns of overlapped triples. Additionally, \cite{wei2020novel} highlighted that NYT and WebNLG datasets are comparatively renowned than the NYT10 and NYT11 datasets. Because NYT10 and NYT11 mainly consist of relation instances belong to the Normal class. They are primarily utilized in the main experiment to show the generalization of our model.

To evaluate the performance of our models, we utilized \emph{Partial Match} and \emph{Exact Match} to measure the performance on the NYT, NYT10, NYT11 and WebNLG datasets. \emph{Partial Match} considers a triple (subject, relation, object) to be correct only if the relation, as well as the tail tokens of both subject and object, is accurate. On the other hand, \emph{Exact Match} deems a correct triple only if the entire span of both entities and the relation are precisely matched.

\textbf{Evaluation Metrics}: in our experiments, we follow the same measurement in previous studies \citep{sui2023joint, ren2022simple, zheng2021prgc}. The standard metrics of micro precision $P$, recall $R$, and $F1$-score are adopted to measure the performance. They are computed as follows:
\begin{equation}
	\begin{aligned}
		P & =\frac{T P}{T P+F P} \times 100 \% \\
		R & =\frac{T P}{T P+F N} \times 100 \% \\
		F 1 & =\frac{2 \times P \times R}{P+R} \times 100 \%
	\end{aligned}
\end{equation}
where $TP$ represents the number of correctly identified positive instances. $FN$ refers to the number of falsely identified negative instances. $FP$ indicates the number of falsely identified positive instances.

\textbf{Implementation Details}: in order to ensure fair comparison, we utilize the Bert-base-cased model \footnote{Available at https://huggingface.co/bert-base-cased}, consisting of 12 Transformer blocks with a hidden size of $D_h = 768$. 
For the WebNLG dataset, a batch size of 6 is set, whereas for other datasets, the batch size is configured to be 5.
We configure the local consolidation module as ‘$[C-A_r-R-V]$’ and optimize all parameters using the Adam \citep{kingma2014adam} with the learning rate of $1 \times 10^{-5}$. 
In Equation (\ref{equ:classifier}), the hidden size of $d = 3 \times D_h$ is utilized for the NYT and $\mathrm{NYT}^*$ dataset, whereas for the remaining datasets, the hidden size of $d = 2 \times D_h$ is employed. Additionally, the maximum sentence length for input is set at 100.

\subsection{Main Results}

In this experiment, our model is compared with twenty one related works. In these models, GraphRel, NovalTagging, CopyRE, ETL-span, and MultiHead utilized LSTM networks as an encoder to generate token representations, while the others applied a pre-trained BERT model to do so. The performance is evaluated in terms of the \emph{Partial Match} and the \emph{Exact Match}. The result is presented in Table \ref{tab:experiment}.

\begin{table*}[h]
	\caption{\label{tab:experiment} Comparing with Related Works. }
	\resizebox{\linewidth}{!}{
		\begin{tabular}{l cccc cccc cccc cccc}
			\toprule[2pt]
			\multirow{3}{*}{Model}  & \multicolumn{8}{c}{$\text{Partial Match}$}  & \multicolumn{8}{c}{$\text{Exact Match}$} \\
			\cline{2-8} \cline{10-16}
			 & \multicolumn{4}{c}{$\mathrm{NYT}^*$}& \multicolumn{4}{c}{$\mathrm{WebNLG}^*$}& \multicolumn{4}{c}{$\mathrm{NYT}$}& \multicolumn{4}{c}{$\mathrm{WebNLG}$}\\
			 \cline{2-4} \cline{6-8} \cline{10-12} \cline{14-16}
			& Prec. & Rec.& F1 &  & Prec.& Rec.& F1 &  & Prec.& Rec.& F1 &  & Prec.& Rec.& F1 & \\
			\midrule[1pt]
			NovalTagging \citep{zheng2017joint}   &-&-&-  &  &-&-&-  &  &32.8&30.6&31.7  &  &52.5&19.3&28.3 & \\
			CopyRE \citep{zeng2018extracting}  &61.0&56.6&58.7  &  &37.7&36.4&37.1  &  &-&-&-  &  &-&-&-  & \\
			MultiHead \citep{bekoulis2018joint}  &-&-&-  &  &-&-&- &  &60.7&58.6&59.6  &  &57.5&54.1&55.7 & \\
			GraphRel \citep{fu2019graphrel}  &63.9&60.0&61.9  &  &44.7&41.1&42.9  &  &-&-&-  &  &-&-&- & \\
			ETL-span \citep{yu2020jointer}  &84.9&72.3&78.1  &  &84.0&91.5&87.6  &  &85.5&71.7&78.0  & &84.3&82.0&83.1  & \\
			\hline
			CasRel \citep{wei2020novel} &89.7&89.5&89.6 & &93.4&90.1&91.8  &  &-&-&-  &  &-&-&-  &\\
			PMEI \citep{sun2021progressive} &90.5&89.8&90.1  &  &91.0&92.9&92.0  & &88.4&88.9&88.7  &  &80.8&82.8&81.8 & \\
			TPLinker \citep{wang2020tplinker} &91.3&92.5&91.9  &  &91.8&92.0&91.9  & &91.4&92.6&92.0  &  &88.9&84.5&86.7 & \\
			StereoRel \citep{tian2021stereorel} &92.0&92.3&92.2  &  &91.6&92.6&92.1  & &92.0&92.3&92.2  &  &-&-&- & \\
			SPN \citep{sui2023joint} &93.3&91.7&92.5  &  &93.1&93.6&93.4  &  &92.5&92.2&92.3  &  &-&-&-  &\\
			PRGC \citep{zheng2021prgc} &93.3&91.9&92.6 & &94.0&92.1&93.0  &  &93.5&91.9&92.7  &  &89.9&87.2&88.5 & \\
			$\text{TDEER}$ \citep{li2021tdeer}  &93.0 &92.1&92.5  &  &93.8&92.4&93.1  &  &-&-&-  &  &-&-&- & \\
			BiRTE \citep{ren2022simple} &92.2&93.8& 93.0 &  &93.2&94.0&93.6 &  &91.9&$\mathbf{93.7}$&92.8  &  &89.0&89.5&89.3 & \\
			RS-TTS \citep{zhang2023rs} &92.9& 92.8 &92.8 & &94.4 &93.9 &94.1  & &93.0&92.6&92.8  &  &90.7 &89.7&90.2 & \\
			CLFM \citep{Wang2023CLFMCL} &93.0 & 92.3 &92.7 & &93.9 &92.6 &93.3  & & 93.3 &92.4 &92.8  &  &90.3 &87.9&89.1 & \\
			BTDM \citep{zhang2023btdm} & 93.0 & 92.5 &92.7 & &94.1 &93.5 &93.8  & &93.1 &92.4 &92.7  &  &90.9 &90.1&90.5 & \\
			OneRel \citep{shang2022onerel} &92.8&92.9&92.8 &  &94.1&94.4&94.3 &  &93.2&92.6&92.9  &  &91.8&90.3&91.0 & \\
			BitCoin \citep{he2023bitcoin} &92.9 & 92.8 &93.1 & &$\underline{94.4}$ & 94.5 & 94.4  & &93.1 &92.6 &92.8  &  &91.9 &90.5&91.2 & \\
			UniRel \citep{tang2022unirel} & $\mathbf{93.5}$& $\mathbf{94.0}$& $\mathbf{93.7}$ &  & $\mathbf{94.8}$&94.6& $\underline{94.7}$ &  &-&-&-  &  &-&-&- & \\
			SOIRP \citep{dai2024soirp} & $\underline{93.3}$& 92.9& 93.1 &  & 94.3 & 94.1 & 94.2 &  & $\underline{93.4}$ & 92.6 & $\underline{93.0}$  &  & $\mathbf{92.7}$ & 89.6 & 91.2 & \\
			RTF \citep{an2024rtf} &93.2 &$\underline{93.4}$&$\underline{93.3}$ &  &94.0&$\underline{95.3}$&94.6 &  &$\mathbf{93.5}$&$\underline{93.2}$&$\mathbf{93.3}$  &  &92.0&$\underline{91.1}$&$\underline{91.6}$ & \\								
			\midrule[1pt]
			Ours  & 92.9 & 93.3& 93.1 &    & 94.1&$\mathbf{96.0}$&$\mathbf{95.0}$ &  &93.1 &92.8& 92.9  & &$\underline{92.2}$&$\mathbf{91.3}$&$\mathbf{91.7}$  & \\
			\midrule[2pt]
			\multirow{3}{*}{Model}  & \multicolumn{8}{c}{$\text{Partial Match}$}  & \multicolumn{8}{c}{$\text{Exact Match}$} \\
			\cline{2-8} \cline{10-16}
			& \multicolumn{4}{c}{$\mathrm{NYT10}$}& \multicolumn{4}{c}{$\mathrm{NYT11}$}& \multicolumn{4}{c}{$\mathrm{NYT10}$}& \multicolumn{4}{c}{$\mathrm{NYT11}$}\\
			\cline{2-4} \cline{6-8} \cline{10-12} \cline{14-16}
			& Prec. & Rec.& F1 &  & Prec.& Rec.& F1 &  & Prec.& Rec.& F1 &  & Prec.& Rec.& F1 & \\
			\midrule[1pt]
			$\text{CasRel}$ \citep{wei2020novel}  &77.7&68.8&73.0  &  &50.1&58.4&53.9  &  &76.8&68.0&72.1  &  &49.1&56.4&52.5 & \\
			$\text{StereoRel}$ \citep{tian2021stereorel}  &80.0&67.4&73.2  &  &53.8&55.4&54.6  &  &-&-&-  &  &-&-&- & \\
			$\text{PMEI}$ \citep{sun2021progressive}  &$\underline{79.1}$&70.4&74.5  &  &$\underline{55.8}$&59.7&57.7  &  &77.3&$\underline{69.7}$&73.3  &  &$\mathbf{54.9}$&58.9&56.8 & \\
			$\text{TPLinker}$ \citep{wang2020tplinker}  &78.9&$\mathbf{71.1}$&$\underline{74.8}$  &  &$\mathbf{55.9}$&$\underline{60.2}$&$\underline{58.0}$  &  &$\underline{78.5}$&68.8&$\underline{73.4}$  &  &\underline{54.8}&$\underline{59.3}$&$\underline{57.0}$ & \\
			\hline
			$\text{Ours}$   &$\mathbf{80.4}$&$\underline{70.8}$&$\mathbf{75.3}$  &  &54.7&$\mathbf{63.2}$&$\mathbf{58.6}$  &  &$\mathbf{79.7}$&$\mathbf{70.3}$&$\mathbf{74.7}$  &  &54.4 &$\mathbf{63.0}$&$\mathbf{58.4}$ & \\
			\bottomrule[2pt]
		\end{tabular}
	}
\end{table*}

As Table \ref{tab:experiment} shown, compared with other models, our model exhibits competitive performance. In the $\mathrm{NYT}^*$ and $\mathrm{WebNLG}^*$ datasets, relation triples are solely annotated with the final word of a given entity. The improvement of our model suggests that the bi-consolidating model is successful in learning global features relevant to a relation instance. On the other hand,  the improvement in the $\mathrm{NYT}$ and $\mathrm{WebNLG}$ also indicates that the bi-consolidating model has the ability to correctly extract the whole span of entities, which heavily depends local features of a sentence. 

In the $\mathrm{NYT}^*$ dataset with $\emph{Partial Match}$, the \emph{UniRel} model has the best performance. The reason may be that the \emph{UniRel} model jointly encodes the representations of entities and relations within a concatenated natural language sequence, which is effective to improve the $\emph{Partial Match}$ measurement. In the NYT dataset, the \emph{RTF} model achieves the best results. The improvement may be attributed to the EPR tagging scheme and the bi-directional decoding algorithm proposed in the \emph{RTF} model. Compared with all related works, our model achieves competitive performance in these datasets. The result demonstrates the effectiveness of our model in supporting relation triple extraction.

At the bottom of Table \ref{tab:experiment}, the NYT10 and NYT11 mainly contain relation instances belong to the Normal class. In comparison to robust model \emph{TPLinker} which is a model based on a local semantic feature that utilizes four different tags for entity and relation detection, our model improves the performance about 0.5\%, 0.6\%, 1.3\%, 1.4\% in F1 scores, respectively. The improvement indicates that our model also has a better generalization capability. Such results confirm the effectiveness of employing the bi-consolidating module to extract all relational triples. It enhances the refinement of semantic features by incorporating both local and global semantic features.

The improvements of the bi-consolidating model are mainly based on three advantages. First, it effectively enhances the local feature information from adjacent regions and mitigate noise in neighbouring neighbours. Secondly, it assimilates relevant information from a global perspective and learn remote semantic dependencies in a sentence, thereby effectively addressing the joint extraction task. Lastly, the utilization of two-dimensional sentence representation gives more semantic information to the sentences and effectively addresses the significant issue of semantic overlapping in relational triple extraction.

\subsection{Ablation Study} 
\label{Ablations of model components}

The purpose of this experiment is to evaluate the contributions of each component in the bi-consolidating model.  Each component is removed individually to evaluate its influence on the final performance. The positional embedding and attention embedding in Equation (\ref{equ:embedding}) are also independently removed to show the influence of sentence encoding. The dataset used for this experiment is the $\mathrm{WebNLG}$. The outcome is presented in Table \ref{table:Ablations of model components}. The influence of each component is discussed as follows.

\begin{table}[!h]
	\small
	\begin{center}
		\caption{\label{table:Ablations of model components} Ablation studies were performed on the WebNLG dataset using the bi-consolidating model under ten distinct settings. }
		\resizebox{\linewidth}{!}{
			\begin{tabular}{l ccc}
				\toprule[2pt]
				Model & Prec. & Rec.  & F1 \\
				\midrule[1pt]
				$\mathbf{Ours}$	&92.2  &91.3  &91.7 \\
				\hline
				\quad w/o Bi-consolidating Module &91.4  &89.0  &90.2   \\
				\quad w/o Global Consolidation  &90.6  &91.1  &90.8 \\
				\quad w/o Local Consolidation  &90.7  &91.4  &91.0 \\
				\quad w/o Spatial Attention   &91.5  &91.3  &91.4 \\ 
				\quad w/o Channel Attention  &91.0  &91.6  &91.3 \\
				\hline
				\quad w/o MaxPooling &91.5  &91.1  &91.3 \\
				\quad w/o AvgPooling &91.6  &90.6  &91.1 \\
				\quad w/o Dimensionality Reduction &91.0  &92.0  &91.5 \\
				\quad w/o Positional Embedding &92.8  &90.2  &91.5 \\
				\quad w/o Attention Embedding &92.5  &90.8  &91.6 \\
				\bottomrule[2pt]    
			\end{tabular}
		}
	\end{center}
\end{table}

(1)  $\emph{w/o Bi-consolidating Module}$ denotes the model removing the \emph{Bi-consolidating Module} from our model. The outcomes demonstrate a substantial decline in performance. It achieves the worst performance in the ablation experiments. Because this model exhibits a dual deficiency, it can not only acquire insight into or extract local pertinent semantic features, but also exhibits limitations in comprehending the comprehensive sentence semantic information from a global point.

(2) $\emph{w/o Global Consolidation}$ denotes the model removing the \emph{Global Consolidation} from the complete model. This model uses only local semantic features to enhance semantic information and loses the advantage of learning distant semantic dependencies in a sentence. Compared to our model, the performance decreases about 0.9\% in the $F1$ score, which confirms the importance of using global semantic features. Our model achieves a significant improvement of approximately 4.1\% in the F1 score compared to the \emph{TPLinker} model. Because \emph{TPLinker} focuses on utilizing local semantic features. Therefore, these findings demonstrate the ability of our model to capture local information. 

(3) $\emph{w/o Local Consolidation}$ denotes the model, which mainly uses global semantic features to enhance the semantic information. In this setting, the F1 score drops about 0.7\% F1 score. The result serves as compelling evidence for the efficacy of the \emph{Global Consolidation} model. This module enhances important features and attenuate unimportant ones controlling the scale, thus rendering the extracted features more directed. Besides, the outcome demonstrates that our model outperforms ``$\emph{w/o Global Consolidation}$'' by a marginal improvement of 0.2\% in terms of the F1 score. It indicates its stronger performance in leveraging global features over local features.

(4) $\emph{w/o Spatial Attention}$ denotes the model excluding the \emph{Spatial Attention Module} from \emph{Global Consolidation}, which results in a less prominent emphasis on salient keywords associated with entities or relations in a given sentence. The F1 score experiences a 0.3\% decrease in comparison to the performance of the complete model, thus affirming the significance of the \emph{Spatial Attention Module}. Spatial attention module allows the model to direct its attention towards salient keywords linked to entities or relationships within a given sentence, thereby reinforcing the significance of the respective word vectors.

(5) $\emph{w/o Channel Attention}$ denotes the model removing the \emph{Channel Attention Module} from \emph{Global Consolidation}. Under this condition, it fails to effectively highlight global semantic features. Our experiments showed that the model performed slightly lower than the “$\emph{w/o Spatial Attention}$”, which indicates that channel attention module plays a vital role in \emph{Global Consolidation}. The channel attention module acts as a core component by simulating the capture of global semantic features. It also indicates that the channel attention is applied globally, whereas the spatial attention operates on a local scale. 

(6) $\emph{w/o MaxPooling}$ denotes the model without the \emph{MaxPooling} operation in \emph{Global Consolidation}. It hinders the model's ability to focus on the most salient features within a given sentence. Our experimental results illustrate that the absence of the \emph{MaxPooling} operation has an adverse impact on both \emph{Spatial Attention Module} and \emph{Channel Attention Module}, ultimately leading to a decline of 0.4\% in the F1 score of the model. This serves as proof that GAP can only acquire sub-optimal features due to the absence of crucial information. However, $\emph{MaxPooling}$ can effectively complement its shortcomings.

(7) $\emph{w/o AvgPooling}$ denotes the model without the \emph{AvgPooling} operation in \emph{Global Consolidation}. Our experiments indicate that this model performs slightly worse than ``$\emph{w/o MaxPooling}$'', implying that average pooling removes some redundant information while maximizing retention of crucial information within a sentence. Therefore, $\emph{AvgPooling}$ represents the most efficient method for preserving the semantic features of a sentence.

(8) $\emph{w/o Dimensionality Reduction}$ denotes the model using the traditional fully-connected (FC) layers with dimensionality reduction in the \emph{Channel Attention Module}. The experimental results illustrate that dimensionality reduction has a negative impact on channel attention prediction, leading to a 0.2\% decrease in the model's F1 score. It explains that fully connected layers without dimension reduction can enhance important features and attenuate unimportant ones controlling the scale, thus rendering the extracted features more directed.

(9) $\emph{w/o Positional Embedding}$: the purpose of positional embedding is to capture the relative positional information between each pair of tokens within a sentence. When the \emph{Positional Embedding} is omitted from the sentence encoding model, the performance decreases, which suggest that the absence of relative location information for each pair of tokens is influential on the final performance. 

(10) $\emph{w/o Attention Embedding}$: the attention embedding is derived from the raw input using an attention layer. This embedding is learned by assigning weights to input features based on their relevance or importance to the output. Experimental results indicate that the presence of \emph{Attention Embedding} positively influences the learning process by refining two-dimensional representations. 

The experimental results show that the performance is significantly declined in the first half of the ablation experiment. On the other hand, components in the second half experiments are less influential. This phenomenon is due to the reason that the first half experiments are relevant to the primary modules, whereas experiments in the second half are sub-operations of primary modules. For example, the positional embedding and attention embedding are sub-components of the two-dimensional semantic representation. MaxPooling, AvgPooling, and Dimensionality Reduction serve as sub-operations in the Global Consolidation module. Therefore, the second half experiments have little impact on the final performance.

\subsection{Performance on Different Overlapping Patterns}

\begin{table*}[h]
	\small
	\caption{\label{tab:results on complex scenarios} Performance on sentences with varying overlapping patterns and varying numbers of triples.  The superscript \S \, indicates that the result is reported by \cite{zheng2021prgc}.}
	\resizebox{0.8\linewidth}{!}{
		\begin{tabular}{c |c| ccccccccc}
			\toprule[2pt]
			Dataset &  Model & Normal & $\mathrm{EPO}$ & $\mathrm{SEO}$ & $\mathrm{SOO}$ & $\mathrm{I}=1$ & $\mathrm{~I}=2$ & $\mathrm{~I}=3$ & $\mathrm{~I}=4$ & $\mathrm{~I} \geq 5$ \\
			\midrule[1pt]
			\multirow{5}{*}{ $\mathrm{NYT}^*$ } & CasRel & 87.3 & 92.0 & 91.4 & $\text{77.0}^{\S}$ & 88.2 & 90.3 & 91.9 & 94.2 & 83.7 \\
			& TPLinker & 90.1 & 94.0 & 93.4 & $\text{90.1}^{\S}$ & 90.0 & 92.8 & 93.1 & 96.1 & 90.0 \\
			& SPN & 90.8 & 94.1 & 94.0 & -  & 90.9 & 93.4 & 94.2 & 95.5 & 90.6 \\
			& PRGC & $91.0$ & 94.5 & 94.0 & 81.8  & 91.1 & 93.0 & 93.5 & 95.5 & 93.0 \\
			& Ours & $\mathbf{91.2}$ & $\mathbf{95.0}$ & $\mathbf{94.9}$ & $\mathbf{92.6}$ & $\mathbf{91.2}$ & $\mathbf{93.7}$ & $\mathbf{93.6}$ & $\mathbf{96.0}$ & $\mathbf{94.3}$  \\
			\hline
			\multirow{5}{*}{ $\mathrm{WebNLG}^*$ } & CasRel & 89.4 & 94.7 & 92.2 & $\text{90.4}^{\S}$ & 89.3 & 90.8 & 94.2 & 92.4 & 90.9 \\
			& TPLinker & 87.9 & 95.3 & 92.5 & $\text{86.0}^{\S}$ & 88.0 & 90.1 & 94.6 & 93.3 & 91.6 \\
			& SPN & 89.5 & 90.8 & 94.1 & -  & 89.5 & 91.3 & 96.4 & 94.7 & 93.8 \\
			& PRGC & 90.4 & 95.9 & 93.6 & 94.6  & 89.9 & 91.6 & 95.0 & 94.8 & 92.8 \\
			& Ours & $\mathbf{91.4}$ & $\mathbf{96.7}$ & $\mathbf{94.9}$ & $\mathbf{96.4}$ & $\mathbf{91.3}$ & $\mathbf{92.7}$ & $\mathbf{96.4}$ & $\mathbf{95.9}$ & $\mathbf{95.4}$  \\
			\bottomrule[2pt]
		\end{tabular}
	}
\end{table*}

This experiment is conducted to validate the effectiveness of the bi-consolidating model in handling overlapping patterns and varying numbers of triples. We utilize four well-established models as comparative baselines. These models are implemented on $\mathrm{NYT}^*$ and $\mathrm{WebNLG}^*$ datasets.

The result in Table \ref{tab:results on complex scenarios} shows that the bi-consolidating model achieves the highest F1-score across all these datasets. Particularly, the performance is improved considerably in $\mathrm{SOO}$ and $I \geq 5$ scenarios, where the overlapping problem is more serious.

The SOO scenario has two distinct characteristics. First, nested entities are difficult to identify, due to their shared contextual features within a sentence. For example, in the triple (``$\emph{Percy Smith}$'', ``$\emph{Family name}$'', ``$\emph{Smith}$''), the object entity is a substring of the subject entity. Second, the difficulty arises when the subject and object contain the same words. For instance, the sentence ``\emph{The ethnic group of native Americans are from the United States where the president is the leader and baked Alaska is a dish}'' in $\mathrm{WebNLG}^*$, which contains a triple (``$\emph{States}$'', ``\emph{ethnicGroup}'', ``$\emph{States}$''). Furthermore, compared with related works, our model shows a significant improvement in $I \geq 5$, where a sentence may hold patterns of SEO, EPO, and SOO all at once. This serves as a clear evidence to show that our model effectively addresses the overlapping triplet problem and its robustness in handling these scenarios.

\subsection{Performance on Different Sub-tasks}

In order to further verify the performance of the bi-consolidating model on the subtasks, we performed experiments on three subtasks in relation triple extraction: $(s, o)$, $r$ and $(s, r, o)$. The $(s, o)$ task denotes to the performance to extract entity pairs. The $r$ task is the performance of relation type classification. Finally, $(s, r, o)$ represents the relation triple extraction task. The $\mathrm{NYT}^*$ and $\mathrm{WebNLG}^*$ datasets are employed in this experiment. Table \ref{table:results on different sub-tasks} presents the performance results.

\begin{table}[h]
	\caption{\label{table:results on different sub-tasks} Performance on sub-tasks of relation triple extraction.}
	\resizebox{0.95\linewidth}{!}{
		\begin{tabular}{l c ccc cccc}
			\toprule[2pt] 
			\multirow{2}{*}{ Model } & \multirow{2}{*}{ Element } & \multicolumn{3}{c}{$\mathrm{NYT}^*$} & \multicolumn{4}{c}{ $\mathrm{WebNLG}^*$} \\
			\cline{3-5}  \cline{7-9}
			& & Prec. & Rec. & F1 & & Prec. & Rec. & F1 \\
			\midrule[1pt]
			\multirow{3}{*}{ CasRel }
			& $(s, o)$ & 89.2 & 90.1 & 89.7 & & 95.3 & 91.7 & 93.5 \\
			& $r$ & 96.0 & 93.8 & 94.9 & & 96.6 & 91.5 & 94.0 \\
			& $(s, r, o)$ & 89.7 & 89.5 & 89.6 & & 93.4 & 90.1 & 91.8 \\
			\midrule[1pt]
			\multirow{3}{*}{ SPN } 
			& $(s, o)$ & 93.2 & 92.7 & 92.9 & & 95.0 & 95.4 & 95.2 \\
			& $r$ & 96.3 & 95.7 & 96.0 & & 95.2 & 95.7 & 95.4 \\
			& $(s, r, o)$ & 93.3 & 91.7 & 92.5 & & 93.1 & 93.6 & 93.4 \\
			\midrule[1pt]
			\multirow{3}{*}{ PRGC } 
			& $(s, o)$ & $\mathbf{94.0}$ & 92.3 & 93.1 & & 96.0 & 93.4 & 94.7 \\
			& $r$ & 95.3 & $\mathbf{96.3}$ & 95.8 & & 92.8 & 96.2 & 94.5 \\
			& $(s, r, o)$ & $\mathbf{93.3}$ & 91.9 & 92.6 & & 94.0 & 92.1 & 93.0 \\
			\midrule[1pt]
			\multirow{3}{*}{ Ours }
			& $(s, o)$ & 93.2 & $\mathbf{94.4}$ & $\mathbf{93.8}$ & & $\mathbf{96.2}$ & $\mathbf{96.6}$ & $\mathbf{96.4}$ \\
			& $r$ & $\mathbf{96.6}$ & 95.6 & $\mathbf{96.1}$ & & $\mathbf{96.2}$ & $\mathbf{97.4}$ & $\mathbf{96.8}$ \\
			& $(s, r, o)$ & 92.9& $\mathbf{93.3}$& $\mathbf{93.1}$ & & $\mathbf{94.1}$& $\mathbf{96.0}$& $\mathbf{95.0}$ \\
			\bottomrule[2pt]
		\end{tabular}
	}
\end{table}

Previous studies \citep{sui2023joint} have shown that the joint extraction task faces two challenges: entity pair recognition and triple formation. For example, suppose a sentence containing three instances: (``$\emph{Macron}$'', ``\emph{President of}'', ``$\emph{France}$''), (``$\emph{Macron}$'', ``\emph{Live in}'', ``$\emph{France}$''), (``$\emph{Macron}$'', ``\emph{Place of birth}'', ``$\emph{France}$''). The recall on relation $r$ may decrease to 0.67 if one relation is misclassified. On the other hand, the recall on entity pair $(s, o)$ may decline to 0 if one entity pair is mistakenly recognized.

The result demonstrates that the bi-consolidating model surpasses all baseline approaches on the three sub-tasks. In the $(s, o)$ task, benefit from the ability of our model to encode local features, we achieve the best performance on recognizing entity pairs. Because the performance of entity pair identification significantly affects the overall results, we also outperform all approaches on the $(s, r, o)$ task. Compared with the $r$ task which only identifies the relation type which relies heavily on global semantic features, our model also has better performance. It further enhances the performance on the $(s, r, o)$ task.

When comparing the features of the $\mathrm{WebNLG}^*$ dataset and the $\mathrm{NYT}^*$ dataset, it is observed that the number of training sets in $\mathrm{WebNLG}^*$ is significantly lower than that in $\mathrm{NYT}^*$ (5019 vs. 56195). Additionally, $\mathrm{WebNLG}^*$ contains a higher number of predefined relations (171 vs. 24) and nearly twice as many SOO patterns as $\mathrm{NYT}^*$ (84 vs. 45).
Therefore, the relation triple extraction task on $\mathrm{WebNLG}^*$ is more challenging than on the $\mathrm{NYT}^*$ task. The result shows that the bi-consolidating model achieves robust performance on the two datasets. The results demonstrate the bi-consolidating model's robust performance on both datasets.

\section{Analysis}
\label{sec:analyis}

In this section, we analyze the influence of several issues on the performance. Firstly, in Section \ref{sec:local_consolidation}, several strategies have been introduced to select pixel pairs from the representation $\mathbf{M}^{so}$. The influence of these selection strategies is analyzed. Secondly, in Section \ref{sec:global_consolidation}, the global consolidation component comprises a channel attention module and a spatial attention module. They can be implemented in a series framework or parallel framework. Thirdly, a deeper analysis of the computational efficiency of the model. Finally, a visualization analysis is given to show more details about our model.

\subsection{Influence of Local Consolidation Architecture}
\label{Configuration of Local Consolidation}

In Figure \ref{fig:pdc}, eight pixel difference convolution operations were presented: CPDC-XY ($C_{xy}$ in short), CPDC-DG ($C_d$), CPDC-OMNI ($C$),  APDC-CW ($A$), APDC-CCW ($A_r$), RPDC-XY ($R_{xy}$), RPDC-DG ($R_d$), RPDC-OMNI ($R$). In these operations, $C_{xy}$ and $C_d$ are sub-operations of C. Except the dependency direction, A and $A_r$ have similar semantic structures. $R_{xy}$ and $R_d$ are also sub-operations of R. After adding the normal convolution operation CNN-2D ($V$), these convolutions can be divided into four groups: [$C_{xy} $, $C_d$, $C$], [$A$, $A_r$], [$R_{xy}$, $R_d$, $R$] and [$V$]. 

A convolution block comprises a pixel difference convolution layer and a normalization layer. The local consolidation component can be stacked with different convolution blocks. Therefore, the architecture of the local consolidation module can be determined by the type of pixel convolutions and the number of blocks. This experiment is conducted to analyse the influence of local consolidation architecture. The performance is evaluated on the $\mathrm{WebNLG}^*$ dataset. Table \ref{table:local_configurations} presents the performance of various architectures, where the top-6 configurations are highlighted in bold.

\begin{table}[h]
	\small
	\caption{\label{table:local_configurations} Influence of convolution architecture.}
	\resizebox{\linewidth}{!}{
		\begin{tabular}{c||c|c|c}
			\toprule[2pt]
			\multicolumn{4}{c}{1 Convolution Block}\\
			\midrule[1pt]
			Architecture & $[C]$ & $[A_r]$ & $[R]$ \\
			P/R/F1 & $93.9/94.8/94.3$ & $93.5/95.4/94.4$ & $93.7/94.5/94.1$ \\
			\midrule[2pt]
			\multicolumn{4}{c}{2 Convolution Blocks} \\
			\midrule[1pt]
			Architecture & $[C_{xy}-C_{d}]$ & $[C-C]$ & $[A-A_r]$ \\
			P/R/F1 & $93.7/95.3/94.5$ & $92.8/95.5/94.2$ & $93.2/95.5/94.3$ \\
			\midrule[1pt]
			Architecture & $[A-A]$ & $[R_{xy}-R_d]$ & $[R-R]$ \\
			P/R/F1 & $93.1/95.4/94.2$ & $93.2/95.5/94.4$ & $94.2/94.3/94.3$ \\
			\midrule[2pt]
			\multicolumn{4}{c}{3 Convolution Blocks}\\
			\midrule[1pt]
			Architecture & $[C-A_r-R]$ & ${[C-A-A_r]}$ & $[C-C-A_r]$ \\
			P/R/F1 & $93.2/95.4/94.3$ & $\mathbf{{93.9}/{95.5}/{94.7}}$ & $93.1/95.2/94.2$ \\
			\midrule[2pt]
			\multicolumn{4}{c}{4 Convolution Blocks}\\
			\midrule[1pt]
			Architecture & ${[C_{xy}-C_{d}] \times 2}$ & $[A-A_r] \times 2$ & $[R_{xy}-R_d] \times 2$ \\
			P/R/F1 & $\mathbf{{94.0}/{95.6}/{94.8}}$ & $93.5/94.9/94.2$ & $93.6/95.4/94.5$ \\
			\midrule[1pt] 
			Architecture & $[C] \times 4$  & $[A_r] \times 4$ & $[R] \times 4$ \\
			P/R/F1 & $93.7/95.2/94.5$  & $93.3/95.6/94.5$ & $93.2/95.4/94.3$ \\
			\midrule[1pt]
			Architecture & $[C_{xy}-C_{d}-A-A_r]$ & $[C_{xy}-C_{d}-R_{xy}-R_d]$  &  $[C_{xy}-C_d-R_d-V]$ \\
			P/R/F1 & $94.0/95.1/94.6$ & $93.8/95.0/94.4$ &  $93.7/95.4/94.5$ \\
			\midrule[1pt]
			Architecture & $[C_{xy}-C_{d}-A_r-R]$ &  $[C_{xy}-C_d-A_r-V]$& $[C_{xy}-A_r-R-V]$ \\
			P/R/F1 & $93.6/95.3/94.5$ &  $93.8/95.1/94.5$ & $93.8/95.3/94.5$ \\
			\midrule[1pt]
			Architecture & $[C-A_r-R_{xy}-V]$  & $[A-A_r-R_{xy}-R_d]$ & $[C-A-R-V]$ \\
			P/R/F1 & $93.6/95.5/94.5$ & $94.1/95.2/94.6$  &  $93.9/95.3/94.6$  \\
			\midrule[1pt]
			Architecture & $[C-A_r-R_{xy}-R_d]$ &  ${[C-A_r-R-R]}$ &   $[C_{d}-A_r-R-V]$\\
			P/R/F1 & $93.9/95.2/94.6$ & $\mathbf{{93.7}/{95.6}/{94.7}}$ &  $\mathbf{93.8}/\mathbf{95.8}/\mathbf{94.8}$   \\
			\midrule[1pt]
			Architecture & $[C-A-A_r-R]$&  $[C-A_r-R_{d}-V]$  & ${[C-A_r-R-V]}$\\
			P/R/F1 &  $94.4/94.8/94.6$ &$\mathbf{93.7/95.7/94.7}$  &   $\mathbf{94.1}$/$\mathbf{96.0}$/$\mathbf{95.0}$\\
			\bottomrule[2pt]
		\end{tabular}}
\end{table}

The results in Table \ref{table:local_configurations} show three fundamental principles to build a local consolidation module.

First, increasing the number of convolution block is helpful to improve the performance. Even stacking identical types of PDCs can provide additional performance gains. The reason is that multi-stacked convolution blocks possess the ability to learn higher-level abstract features, which are beneficial in supporting the task of relation triple extraction. Furthermore, stacked blocks can enlarge the receipt field of the input, making it effective for learning semantic dependencies within a sentence.

Second, stacking convolution blocks with different pixel difference operations can achieve better performance. For example, the two configurations $[C_{d}-A_r-R-V]$ and ${[C-A_r-R-V]}$ have the best performance, which utilize various types of convolutions for capturing semantic information in different encoding directions. But there are exceptions. For example, the bidirectional APDCs $[C-A-A_r]$ also demonstrates the superiority of local consolidations, which may leverage semantic contextual information from adjacent elements to aid in entity boundary recognition. $[C_{xy}-C_d] \times 2$ not only underscores its prioritization of extracting features from central and adjacent positions, but also showcases the substantiated efficiency of acquiring information from both the XY axis and diagonal directions. 

Third, replacing pixel difference convolutions in the same group is less influential on the final performance, e.g., [$C_{xy} $, $C_d$, $C$], [$A$, $A_r$], [$R_{xy}$, $R_d$, $R$]. The reason may be that, in the same group, convolution operations capture semantic structures in the same pattern. Furthermore, in the same group, the OMNI-version operations (e.g., $C$ and $R$) usually have better performance, because other pixel difference convolutions in the same group are usually sub-operations of the OMNI-version operation.

\subsection{Influence of Global Consolidation Architecture}

The global consolidation component contains a channel attention module and a spatial attention module, which can be set in a series or parallel framework. To show the influence of the global consolidation architecture, in this experiment, we evaluate the model's performance by organizing them in a series or parallel framework. In the series framework, the output of the sentence encoding module is sequentially processed by the channel attention module and spatial attention module, respectively. In the parallel framework, the channel attention and spatial attention are paralleled. Furthermore, in the whole bi-consolidating module, the local consolidation component and the global consolidation component can be placed into a series or a parallel framework too. The influence is also evaluated in this section.

In this experiment, the $\mathrm{WebNLG^*}$ dataset is employed to evaluate the performance. The influence of the architectures is listed in Table \ref{Series and Parallel Configurations}, where the symbol “\&” signifies a parallel connection, while “$+$” is employed to denote a series connection.

\begin{table}[!h]
	\begin{center}
		\caption{\label{Series and Parallel Configurations} Performance with different global consolidation architecture.}
		\resizebox{0.8\linewidth}{!}{
			\begin{tabular}{l ccc}
				\toprule[2pt]
				Model & Prec. & Rec.  & F1 \\
				\midrule[1pt]
				\multicolumn{4}{l}{Within Global Consolidation} \\
				\quad Channel \& Spatial in parallel &$\mathbf{94.5}$	&95.3	&94.9   \\
				\quad Channel $+$ Spatial	 &94.1	&$\mathbf{96.0}$	&$\mathbf{95.0}$    \\
				\quad Spatial $+$ Channel	&93.5	&95.8	&94.6     \\
				\midrule[1pt]
				\multicolumn{4}{l}{Within Bi-consolidating Module}  \\
				\quad Local \& Global in parallel  &93.7	&95.7	&94.7 \\
				\quad Local $+$ Global  &$\mathbf{94.1}$	&$\mathbf{96.0}$	&$\mathbf{95.0}$  \\
				\bottomrule[2pt]    
			\end{tabular}
		}
	\end{center}
\end{table}

Table \ref{Series and Parallel Configurations} presents the experimental results of different methods for arranging modules, which indicates that arranging the channel attention and spatial attention in a series framework leads to better performance compared to parallel framework. The reason is that sequentially processing the input has the advantage to learn the interaction between channel and spatial semantic space. Moreover, the channel-prior order demonstrates superior performance over the spatial-prior order. This trend can be seen in the \emph{Bi-consolidating Module} too, where sequentially generating a refined subject-object features also yields better results than implementing them in parallel.

\subsection{Computational Efficiency}

The bi-consolidating model is mainly composed of a global consolidation component and a local consolidation component. The global consolidation component contains a channel attention module and a spatial attention module, which are normal operations in natural language processing. The local consolidation component is stacked of four convolution blocks, where different pixel difference convolutions are used for learning high-order abstract features. In order to evaluate the computational efficiency of the bi-consolidating model, we compare the training times between five architectures with different local consolidation components: $[C-A_r-R-V]$, $[V-V-V-V]$, $[V-A_r-R-V]$, $[C-V-R-V]$ and $[C-A_r-V-V]$.

\begin{figure}[h] 
	\centering
	\includegraphics[width=\linewidth]{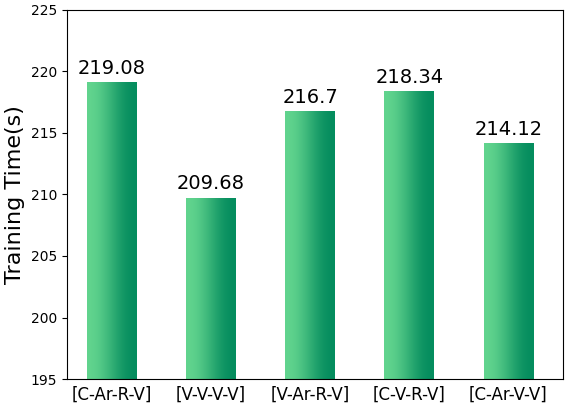}
	\caption{Computational efficiency.}
	\label{fig:Computational Efficiency}
\end{figure}

In this experiment, for making a fair comparison, instead of directly compared with a deep architecture composed of a convolution layer, the $[V-V-V-V]$  model is used as the baseline, which is composed of four normal convolution layers. This experiment is conducted on the $\mathrm{WebNLG^*}$ dataset. The batch size for training and testing are set to 6 and 1, respectively. The max length of input sentence is 100. The result is shown in Figure \ref{fig:Computational Efficiency}, where the training time (second) means the time required to implement one epoch in the training process.

In Figure \ref{fig:Computational Efficiency}, the four architectures achieve similar performance. This result indicates that different pixel convolution operations are less influential on the computational complexity. Compared the performance between them, $[V-V-V-V]$ achieves the lowest computational complexity. The reason is that  $[V-V-V-V]$ only implements convolution operation. It is effective to support parallel computing. On the other hand, as shown in Figure \ref{fig:pdc}, the $C$, $R$, $A_r$ and $A$ convolution operations need to compute the semantic dependencies of elements relevant to adjacent elements, which limit the utilization of parallel computing. Especially, the $R$ operation has a larger kernel size, which further degenerates the final performance. Even so, compared $[C-A_r-V-V]$ with $[V-V-V-V]$,  it only decreases about 5\% computing speed.

\subsection{Visualization Of Different Bi-consolidating Module Components}

\begin{figure*}[h]
	\centering
	\includegraphics[width=\linewidth]{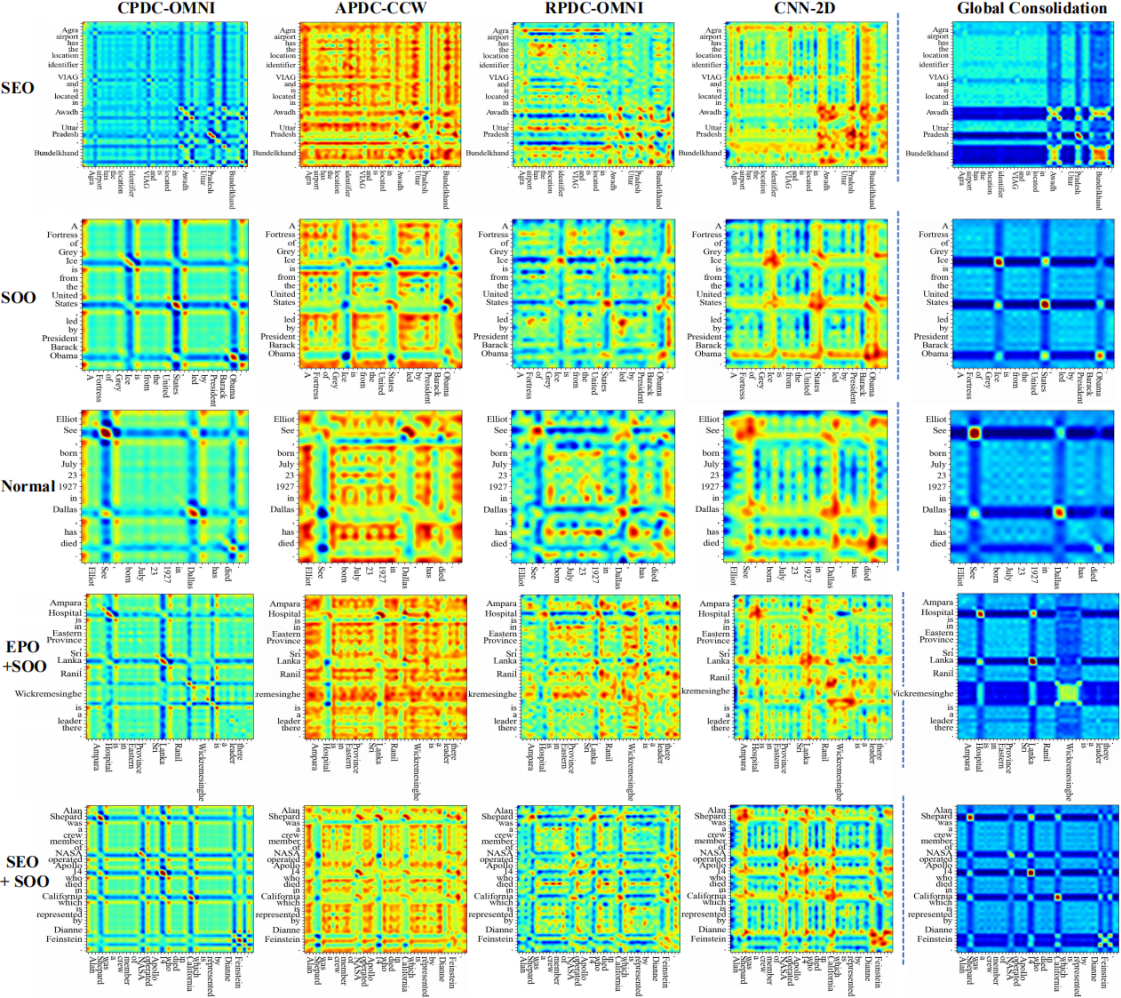}
	\caption{\label{fig:visual semantic overlap} The visualization of semantic overlap surrounding the entities of various components of the bi-consolidating module.}
\end{figure*}

A 2D sentence representation unfolds a two-dimensional semantic space. This approach effectively expresses the semantic structures of a sentence and provides additional semantic information. The 2D representation is fed into the bi-consolidating module to generate a fine-grained subject-object features, encoded with local and global semantic features. In addition to quantitative analysis, it is also informative to provide a visualized analysis to show the influence of the bi-consolidating module. In this visualization, we collect five sentences with different overlapping patterns as SEO, SOO, Normal, EPO + SOO, SEO + SOO. These sentences are collected from the $\mathrm{WebNLG^*}$ dataset. They are introduced as follows:

1) SEO: the sentence is ``\emph{Agra airport has the location identifier VIAG and is located in Awadh, Uttar Pradesh, Bundelkhand}''. It contains two relational triples as: [``\emph{Pradesh}'', ``\emph{isPartOf}'', ``\emph{Awadh}''], [``\emph{Pradesh}'', ``\emph{isPartOf}'', ``\emph{Bundelkhand}'']. The entity pairs (``\emph{Pradesh}'', ``\emph{Awadh}'') and (``\emph{Pradesh}'', ``\emph{Bundelkhand}''), with the overlapping subject ``\emph{Pradesh}'', belong to a single entity overlap.

2) SOO: the sentence is ``\emph{A Fortress of Grey Ice is from the United States, led by President Barack Obama}''. It contains two relational triple as: [``\emph{States}'', ``\emph{leaderName}'', ``\emph{Obama}''], [``\emph{Ice}'', ``\emph{country}'', ``\emph{States}'']. The entity pairs (``\emph{States}'', ``\emph{Obama}'') and (``\emph{Ice}'', ``\emph{States}'') have the entity ``\emph{States}'' as their subject and object, respectively, resulting in a subject object overlap.

3) Normal: the sentence is ``\emph{Elliot See, born July 23 1927 in Dallas, has died}''. It contains a single relational triple as: [``\emph{See}'', ``\emph{birthPlace}'', ``\emph{Dallas}'']. There is no overlap among them.

4) EPO + SOO: the sentence is ``\emph{Ampara Hospital is in Eastern Province, Sri Lanka. Ranil Wickremesinghe is a leader there}''. It contains three relational triples as: [``\emph{Hospital}'', ``\emph{country}'', ``\emph{Lanka}''], [``\emph{Lanka}'', ``\emph{leaderName}'', ``\emph{Wickremesinghe}''] and [``\emph{Hospital}'', ``\emph{state}'', ``\emph{Lanka}'']. The entity pairs (``\emph{Hospital}'', ``\emph{Lanka}'') and (``\emph{Lanka}'', ``\emph{Wickremesinghe}'') have the entity ``\emph{Lanka}'' as their subject and object, resulting in a subject object overlap.

5) SEO + SOO: the sentence is ``\emph{Alan Shepard was a crew member of NASA operated Apollo 14 who died in California which is represented by Dianne Feinstein}''. It contains four relational triples as: [``\emph{Shepard}'', ``\emph{was a crew member of}'', ``\emph{14}''], [``\emph{Shepard}'', ``\emph{deathPlace}'', ``\emph{California}''], [``\emph{California}'', \emph{senators}, ``\emph{Feinstein}''] and [``\emph{14}'', ``\emph{operator}'', ``\emph{NASA}'']. The entity pairs (``\emph{Shepard}'', ``\emph{14}'') and (``\emph{Shepard}'', ``\emph{California}'') belong to a single entity overlap. The entity pairs (``\emph{Shepard}'', ``\emph{California}'') and (``\emph{California}'', ``\emph{Feinstein}'') have the entity ``\emph{California}'' as their subject and object, resulting in a subject object overlap.

These sentences are first fed into the sentence encoding module to generate $\mathbf{M}^{so} \in \mathbb{R}^{N \times N \times D_h}$. Then, the $\mathbf{M}^{so}$ is fed into the bi-consolidating module to encode the subject and object features of all possible entity pairs in a sentence. In our model, the bi-consolidating module is composed of a local consolidation component and a global consolidation component. The local consolidation contains 4 convolution blocks referred as: ``CPDC-OMNI'', ``APDC-CCW'', ``RPDC-OMNI'' and ``CNN-2D''. To show the visualization of various components in the bi-consolidating module, we transform each 2D representation an image by implementing the Global Average Pooling operation. This resulting image can be seen as a heat map, which denotes to a semantic plane. The visualization of these semantic planes in the five sentences is shown in Figure \ref{fig:visual semantic overlap}.

Figure \ref{fig:visual semantic overlap} is divided into five parts marked with ``SEO'', ``SOO'', ``Normal'', ``EPO + SOO'' and ``SEO + SOO''. Each part is composed of five images belonging to a sentence, which are further categorized into local consolidation section and global consolidation section. The local consolidation section shows the visualization of each block as ``CPDC-OMNI'', ``APDC-CCW'', ``RPDC-OMNI'', ``CNN-2D''. 

It is interesting to observe that after the utilization of the convolution ``CPDC-OMNI'', all entities exhibit distinct contour characteristics. For instance, in the ``Normal'', the entities ``See'' and ``Dallas'' have a high intensity. Their span edges exhibit significant semantic intensity changes in the semantic plane. Next, following the convolution block ``APDC-CCW'', the semantic representation surrounding the entity is amplified, which guarantees the retention of all the semantic information within the sentence. Next to the convolution block ``RPDC-OMNI'', the semantic information within a neighborhood is intensified, which mitigates any noise that might exist in adjacent elements. Subsequently, after the convolution block ``CNN-2D'', the contextual semantic representation in the proximity of the entity undergoes further enhancement, simultaneously minimizing potential disturbances originating from neighboring elements. 

As visualized in every part, it can be seen that all entities exhibit greater intensity. On the other hand, non-entity regions manifest comparatively lower intensity. In the visualization of the global consolidation module, the most impressive phenomenon is that all entity semantic representations have clear contour characteristics and have a high intensity in a 2D sentence representation. This also proves that this module has the advantage to learn remote semantic dependencies in a sentence. Another interesting phenomenon is that representations, composed of the semantic information about an entity pair, also have concretized semantic representations.

\section{Conclusion}
\label{sec:conclusion}

In this paper, based on a two-dimensional sentence representation, we propose a bi-consolidating model to address the semantic overlapping problem. Our model consists of a local consolidation component and a global consolidation component.
The first component uses a pixel difference convolution to enhance semantic information of a possible triple representation of adjacent regions and mitigate noise in neighbouring neighbours.
The second component strengthens the triple representation based a channel attention and a spatial attention, which has the advantage to learn remote semantic dependencies in a sentence. The bi-consolidating model simultaneously reinforce the local and global semantic features relevant to each relation triple.
Our experiments demonstrate that on many benchmark datasets, the bi-consolidating model competitive performance.  
In-depth analyses are carried out to comprehend the influence of the components of the bi-consolidating model on performance and to validate their significance.
Future research can focus on developing this model as a generalized framework to support various related multi-objective NLP tasks.

\section*{Acknowledgements}

This work is supported by National Key R\&D Program of China under Grant No. 2023YFC3304500, the Major Science and Technology Projects of Guizhou Province under Grant [2024]003 and National Natural Science Foundation of China under Grant No. 62066008 and No. 62066007.


\bibliographystyle{cas-model2-names}

\bibliography{cas-refs}

\hspace*{\fill} 
\subsection*{  } 
\setlength\intextsep{0pt} 
\begin{wrapfigure}{l}{25mm}
	\centering
	\includegraphics[width=1in,height=1.25in,clip,keepaspectratio]{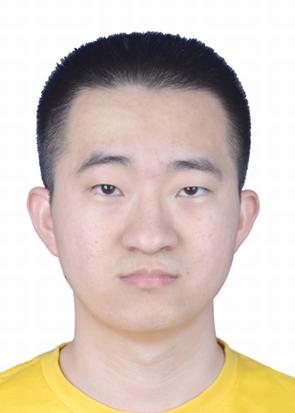}
\end{wrapfigure}
\noindent \textbf{Xiaocheng Luo} He is currently pursuing his master’s degree in Computer Science and Technology at Guizhou University. His research interests include Information extraction. \par

\hspace*{\fill} 
\subsection*{  } 
\setlength\intextsep{0pt} 
\begin{wrapfigure}{l}{25mm}
	\centering
	\includegraphics[width=1in,height=1.25in,clip,keepaspectratio]{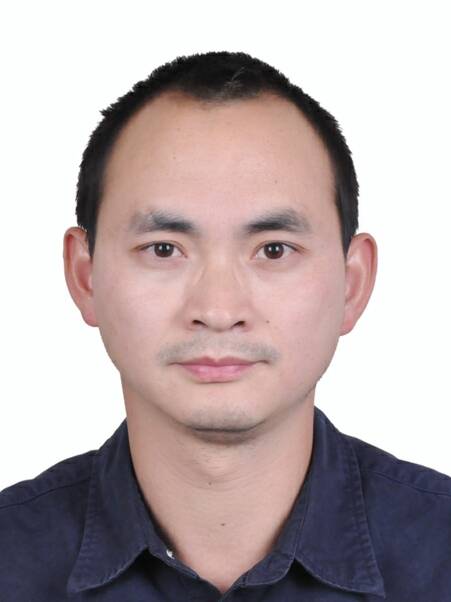}
\end{wrapfigure}
\noindent \textbf{Yanping Chen} He is a full professor at the School of Computer Science and Technology, Guizhou University. His research interests include natural language processing, information extraction. \par

\hspace*{\fill} 

\subsection*{  } 
\setlength\intextsep{0pt} 
\begin{wrapfigure}{l}{25mm}
	\centering
	\includegraphics[width=1in,height=1.25in,clip,keepaspectratio]{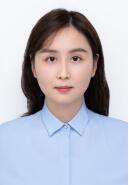}
\end{wrapfigure}
\noindent \textbf{Ruixue Tang} She is a full professor at the School of Computer Science and Technology, Guizhou University of Finance and Economics. Her research interests include Natural language processing, information extraction and machine learning.  \par


\subsection*{  } 
\setlength\intextsep{0pt} 
\begin{wrapfigure}{l}{25mm}
	\centering
	\includegraphics[width=1in,height=1.25in,clip,keepaspectratio]{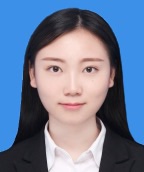}
\end{wrapfigure}
\noindent \textbf{Caiwei Yang} She is currently pursuing her master's degree at Guizhou University, China. Her current research interests are in the area of natural language processing and information extraction.  \par

\subsection*{  } 
\subsection*{  } 
\setlength\intextsep{0pt} 
\begin{wrapfigure}{l}{25mm}
	\centering
	\includegraphics[width=1in,height=1.25in,clip,keepaspectratio]{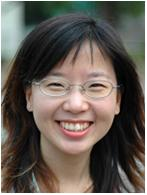}
\end{wrapfigure}
\noindent \textbf{Ruizhang Huang} She is a full professor in the School of Computer Science and Technology, Guizhou University. Her research interests include text mining, machine learning and web mining. \par

\hspace*{\fill} 

\subsection*{  } 
\setlength\intextsep{0pt} 
\begin{wrapfigure}{l}{25mm}
	\centering
	\includegraphics[width=1in,height=1.25in,clip,keepaspectratio]{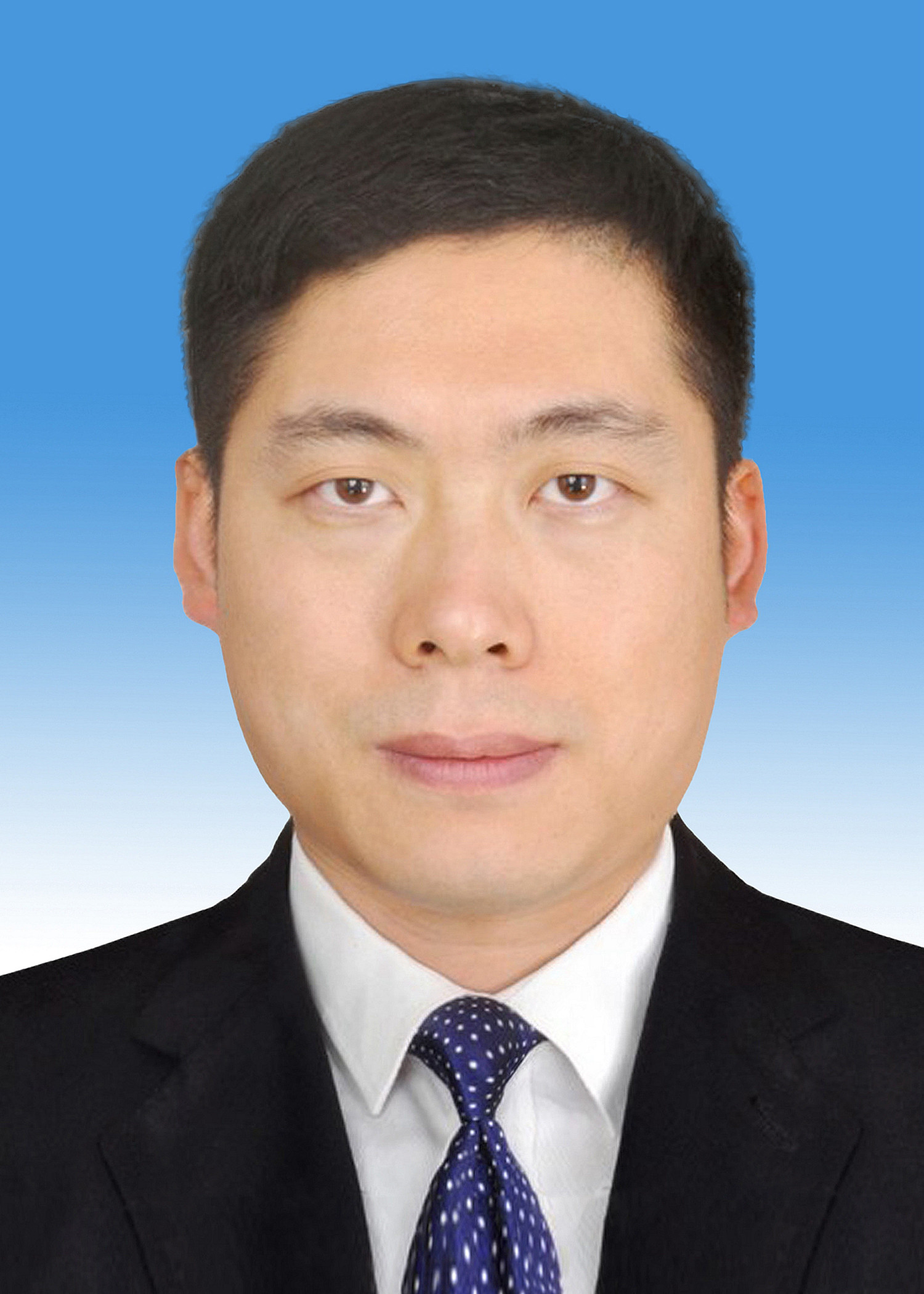}
\end{wrapfigure}
\noindent \textbf{Yongbin Qin} He is a full professor at the School of Computer Science and Technology, Guizhou University. His research interests include text analytics, big data governance and application, enterprise informatization and e-government. \par

\end{document}